\documentclass[11pt,a4paper]{article}
\usepackage[hyperref]{emnlp2020}
\usepackage{times}
\usepackage{latexsym}

\usepackage{microtype}

\usepackage{
  amsfonts, nicefrac, amsmath, bm, hyperref,
  soul, graphicx, enumitem, multirow, wrapfig, lipsum, booktabs, subcaption, stfloats,%
}
\usepackage[frozencache,cachedir=.]{minted}
\usepackage[capitalise]{cleveref}
\usepackage{longtable}

\newcommand{\ho}[1]{\textcolor{blue}{}}
\newcommand{\pg}[1]{\textcolor{red}{}}
\newcommand{\psrcomment}[1]{}

\newcommand{\ignore}[1]{}
\newcommand{\ourmodel}{BAT }

\newcommand{\B}{\textbf{B}}

\aclfinalcopy %

\title{Improving Neural Topic Models using Knowledge Distillation}

\author{Alexander Hoyle\thanks{\, Equal contribution.} \\
  Computer Science \\
  University of Maryland \\
  College Park, MD \\
  \texttt{hoyle@umd.edu} \\\And
  Pranav Goel\footnotemark[1] \\
  Computer Science \\
  University of Maryland \\
  College Park, MD \\
  \texttt{pgoel1@umd.edu} \\\And
  Philip Resnik \\
  Linguistics / UMIACS \\
  University of Maryland \\
  College Park, MD \\
  \texttt{resnik@umd.edu} \\}

\date{}

\begin{document}
\maketitle

\begin{abstract}

  Topic models are often used to identify human-interpretable topics to help make sense of large document collections. We use knowledge distillation to combine the best attributes of probabilistic topic models and pretrained transformers. Our modular method can be straightforwardly applied with any neural topic model to improve topic quality, which we demonstrate using two models having disparate architectures, obtaining state-of-the-art topic coherence. We show that our adaptable framework not only improves performance in the aggregate over all estimated topics, as is commonly reported, but also in head-to-head comparisons of aligned topics.

\end{abstract}

\section{Introduction}

\noindent The core idea behind the predominant \emph{pretrain and fine-tune} paradigm for transfer learning in NLP is that general language knowledge, gleaned from large quantities of data using unsupervised objectives, can serve as a foundation for more specialized endeavors.
Current practice involves taking the full model that has amassed such general knowledge and fine-tuning it with a second objective appropriate to the new task \citep[see][for an overview]{raffelExploringLimitsTransfer2019}.
Using these methods, pre-trained transformer-based language models \citep[e.g., BERT, ][]{devlin-etal-2019-bert} have been employed to great effect on a wide variety of NLP problems, thanks, in part, to a fine-grained ability to capture aspects of linguistic context \citep{clark-etal-2019-bert,liu-etal-2019-linguistic,rogersPrimerBERTologyWhat2020}.

\begin{figure}[ht!]
  \centering
  \includegraphics[width=1.\columnwidth,
    keepaspectratio]{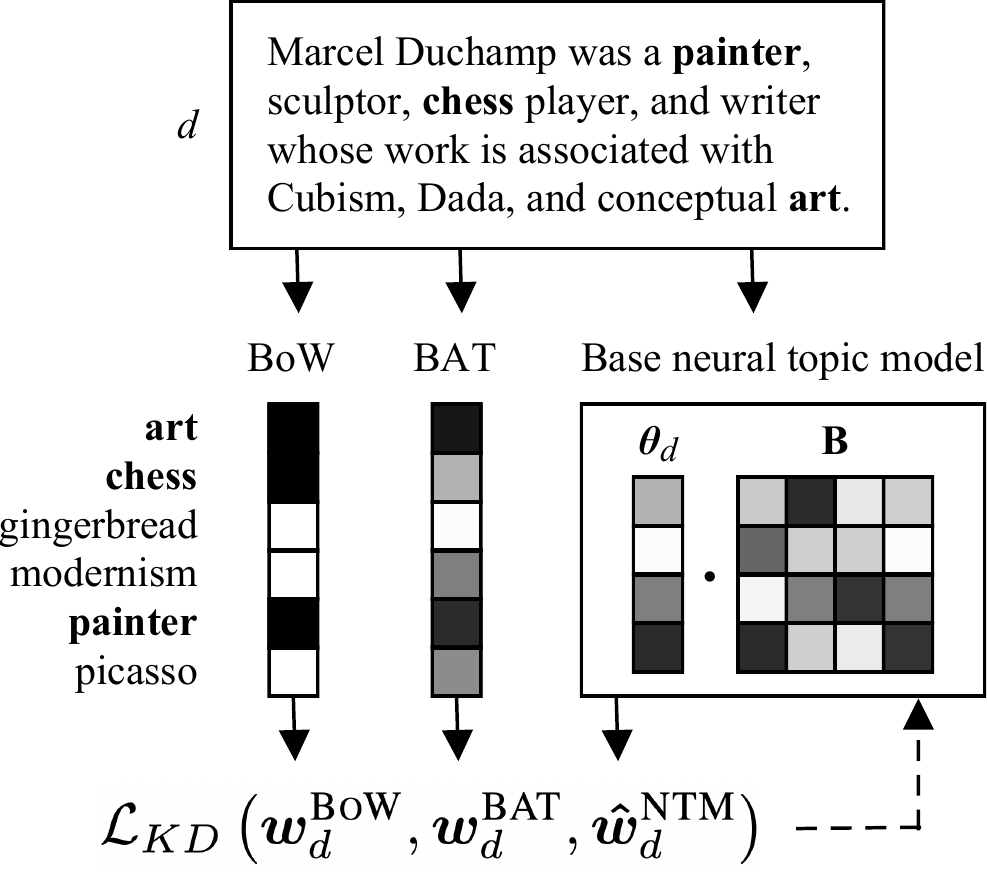}
  \caption{Improving a base neural topic model with knowledge distillation. A document is mapped through both a standard BoW representation and a \textbf{B}ERT-based \textbf{A}uto-encoder ``\textbf{T}eacher'' (\textsc{BAT}), yielding two distributions over words. These are used as the ground truth in the ``student'' topic model's document reconstruction loss $\mathcal{L}_{KD}$ (backpropagated along the dotted line). Crucially, the \ourmodel distribution assigns mass to unobserved but related terms (unbolded). 
    }
  \label{fig:thresh_matched_topics}
\end{figure}

However, this paradigm introduces a subtle but insidious limitation that becomes evident when the downstream application is a topic model.
A topic model may be cast as a (stochastic) autoencoder \citep{miaoNeuralVariationalInference2016}, and we could fine-tune a pretrained transformer with an identical document reconstruction objective. But in replacing the original topic model, we lose the property that makes it desirable: its interpretability. The transformer gains its contextual power from its ability to exploit a huge number of parameters, while the interpretability of a topic model comes from a dramatic dimensionality reduction.

We combine the advantages of these two approaches---the rich contextual language knowledge in pretrained transformers and the intelligibility of topic models---using \emph{knowledge distillation} \citep{hintonDistillingKnowledgeNeural2015}. In the original formulation, knowledge distillation involves training a parameter-rich \emph{teacher} classifier on large swaths of data, then using its high-quality probability estimates over outputs to guide a smaller \emph{student} model.
Since the information contained in these estimates is useful---a picture of an ox will yield higher label probabilities for \textsc{buffalo} than \textsc{apricot}---the student needs less data to train and can generalize better.

We show how this principle can apply equally well to improve unsupervised topic modeling, which to our knowledge has not previously been attempted.  While distillation usually involves two models of the same type, it \emph{can} also apply to models of differing architectures.
Our method is conceptually quite straightforward: we fine-tune a pretrained transformer \citep{sanhDistilBERTDistilledVersion2019} on a document reconstruction objective, where it acts in the capacity of an autoencoder. When a document is passed through this BERT autoencoder, it generates a distribution over words that includes unobserved but related terms. We then incorporate this distilled document representation into the loss function for topic model estimation. (See Figure~\ref{fig:thresh_matched_topics}.) 

To connect this method to the more standard supervised knowledge distillation, observe that the unsupervised ``task'' for both an autoencoder and a topic model is the reconstruction of the original document, i.e. prediction of a distribution over the vocabulary. The BERT autoencoder, as ``teacher'', provides a dense prediction that is richly informed by training on a large corpus. The topic model, as ``student'', is generating its own prediction of that distribution. We use the former to guide the latter, essentially as if predicting word distributions were a multi-class labeling problem.\footnote{An interesting conceptual link here can be found in Latent Semantic Analysis \citep[LSA, ][]{landauerSolutionPlatoProblem1997}, an early predecessor of today's topic models. The original discussion introducing LSA has a very autoencoder-like flavor, explicitly illustrating the deconstruction of a collection of sparsely represented documents and the reconstruction of a dense document-word matrix.}
Our approach, which we call \textbf{B}ERT-based \textbf{A}utoencoder as \textbf{T}eacher \textbf{(BAT)}, obtains best-in-class results on the most commonly used measure of topic coherence, normalized pointwise mutual information \citep[NPMI, ][]{aletras-stevenson-2013-evaluating} compared against recent

state-of-the-art-models that serve as our baselines. 

In order to accomplish this, we adopt neural topic models \citep[NTM,][\emph{inter alia}]{miaoNeuralVariationalInference2016,Srivastava2017AutoencodingVI, card-etal-2018-neural, burkhardtDecouplingSparsitySmoothness2019, nan-etal-2019-topic}, which use various forms of black-box distribution-matching \citep{kingmaAutoEncodingVariationalBayes2014,tolstikhinWassersteinAutoEncoders2018}.\footnote{As a standard example, \newcite{Srivastava2017AutoencodingVI} encode a document's bag-of-words with a neural network to parameterize the latent topic distribution, then sample from the distribution to reconstruct the BoW.}
These now surpass traditional methods
\citep[e.g. LDA, ][and variants]{bleiLatentDirichletAllocation2003} in topic coherence. In addition, it is easier to modify the generative model of a neural topic model than for a
classic probabilistic latent-variable model, where changes generally require investing effort in new variational inference procedures or samplers. In fact, because we leave the base NTM unmodified, our approach is flexible enough to easily accommodate \emph{any} neural topic model, so long as it includes a word-level document reconstruction objective.
We support this claim by demonstrating improvements on models based on both Variational \citep{card-etal-2018-neural} and Wasserstein \citep{nan-etal-2019-topic} auto-encoders.

To summarize our contributions:
\begin{itemize}
  \item We introduce a novel coupling of the knowledge distillation technique with generative graphical models.
  \item We construct knowledge-distilled neural topic models that achieve better topic coherence than their counterparts without distillation on three standard English-language topic-modeling datasets.
  \item We demonstrate that our method is not only effective but modular, by improving topic coherence in a base state-of-the-art model by modifying only a few lines of code.\footnote{See \cref{appendix:code}. Our full implementation, including dataset preprocessing, is available at \url{github.com/ahoho/kd-topic-models}.}
        
  \item In addition to showing overall improvement across topics,  our method preserves the topic analysis of the base model and improves coherence on a topic-by-topic basis.\looseness=-1
\end{itemize}

\section{Methodology}\label{sec:background}

\subsection{Background on topic models}\label{sec:background:topicmodels}

Topic modeling is a well-established probabilistic method that aims to summarize large document corpora using a much smaller number of latent \emph{topics}. The most prominent instantiation, LDA \citep{bleiLatentDirichletAllocation2003}, treats each document as a mixture over $K$ latent topics, $\bm{\theta}_d$, where each topic is a distribution over words $\bm{\beta}_k$. By presenting topics as ranked word lists and documents in terms of their probable topics, topic models can provide legible and concise representations of both the entire corpus and individual documents.

In classical topic models like LDA, distributions over the latent variables are estimated with approximate inference algorithms tailored to the generative process. Changes to the model specification---for instance, the inclusion of a supervised label---requires attendant changes in the inference method, which can prove onerous to derive. For some probabilistic models, this problem may be circumvented by the variational auto-encoder \cite[VAE,][]{kingmaAutoEncodingVariationalBayes2014}, which introduces a \emph{recognition model} that approximates the posterior with a neural network. As a result, \emph{neural topic models} have capitalized on the VAE framework \cite[][\emph{inter alia}]{Srivastava2017AutoencodingVI,card-etal-2018-neural,burkhardtDecouplingSparsitySmoothness2019} and other deep generative models \citep{wangATMAdversarialneuralTopic2019,nan-etal-2019-topic}. In addition to their flexibility, the best models now yield more coherent topics than LDA.

Although our method (\cref{sec:methodology}) is agnostic as to the choice of neural topic model, we borrow from \citet{card-etal-2018-neural} for both formal exposition and our base implementation (\cref{sec:experiment_setup}). \citet{card-etal-2018-neural} develop \textsc{Scholar}, a generalization of the first successful VAE-based neural topic model \citep[\textsc{ProdLDA},][]{Srivastava2017AutoencodingVI}. The generative story is broadly similar to that of LDA, although the uniform Dirichlet prior is replaced with a logistic normal ($\mathcal{LN}$):\footnote{This choice is because the reparameterization trick behind VAEs used to be limited to location-scale distributions, but recent developments \citep[e.g.,][]{Figurnov2018ImplicitRG} have lifted that restriction, as \citet{burkhardtDecouplingSparsitySmoothness2019} demonstrate with several Dirichlet-based NTMs using VAEs.}
\begin{itemize}
    \item[] For each document $d$:
          \begin{itemize}\vspace{-0.3em}
              \item Draw topic distribution $\bm{\theta}_d \sim \mathcal{LN}(\bm{\alpha}_0)$
              \item For each word $w_{id}$ in the document:
                    \begin{itemize}
                        \item[] $w_{id} \sim \text{Multinomial}\left(1, f(\bm{\theta}_d, \B)\right)$
                    \end{itemize}
          \end{itemize}
\end{itemize}
Following \textsc{ProdLDA}, $\B$ is a $K \times V$ matrix where each row corresponds to the $k$th topic-word probabilities in log-frequency space. The multinomial distribution over a document's words is parameterized by
\begin{align}
    f(\bm{\theta}_d, \B) = \sigma\left( \bm{m} + \bm{\theta}_d^\top \B \right)\label{eq:word-prob}
\end{align}
where $\bm{m}$ is a vector of fixed empirical background word frequencies and $\sigma(\cdot)$ is the softmax function. We highlight that each document is treated as a bag of words, $\bm{w}^\textsc{BoW}_d$.

To perform inference on the model, VAE-based models like $\textsc{Scholar}$ approximate the true intractable posterior $p(\bm{\theta}_d \mid \cdot)$ with a neural \emph{encoder} network $g(\bm{w}_d)$ that parameterizes the variational distribution $q\left(\bm{\theta}_d \mid g(\cdot)\right)$ (here, a logistic normal with diagonal covariance). The Evidence Lower BOund (ELBO) is therefore
\begin{align}
    \text{ELBO}   & = \mathbb{E}_{q(\bm{\theta}_d \mid\, \cdot \,)}\big[\,\mathcal{L}_R\,\big]\nonumber                                          \\
                  & -\text{KL}\big[ q\left(\bm{\theta}_d \mid \bm{w}^\textsc{BoW}_d, \bm{x}_d\right) \mid\mid p\left(\bm{\theta}_d\right) \big], \\
    \mathcal{L}_R & = \left(\bm{w}_d^{\textsc{BoW}}\right)^\top \log f(\bm{\theta}_d, \B), \label{eq:elbo-recon}
\end{align}
which is optimized with stochastic gradient descent. The form of the reconstruction error $\mathcal{L}_R$ is a consequence of the independent multinomial draws.\looseness=-1

\subsection{Background on knowledge distillation}\label{sec:background:kd}
It is instructive to think of \cref{eq:word-prob} as a latent logistic regression, intended to approximate the distribution over words in a document. Under this lens, the neural topic model outlined above can be cast as a multi-label classification problem. Indeed, it accords with the standard structure: there is a softmax over logits estimated by a neural network, coupled with a cross-entropy loss.

However, because $\bm{w}^\textsc{BoW}_d$ is a sparse bag of words, the model is limited in its ability to generalize. During backpropagation (\cref{eq:elbo-recon}), the topic parameters will only update to account for observed terms, which can lead to overfitting and topics with suboptimal coherence.

In contrast, dense document representations can capture rich information that bag-of-words representations cannot.

These observations motivate our use of \emph{knowledge distillation} \citep[KD,][]{hintonDistillingKnowledgeNeural2015}. The authors argue that the knowledge learned by a large ``cumbersome'' classifier on extensive data---e.g., a deep neural network or an ensemble---is expressed in its probability estimates over classes, and not just contained in its parameters. Hence, these teacher estimates for an input may be repurposed as soft labels to train a smaller student model. In practice, the loss against the true labels is linearly interpolated with a loss against the teacher probabilities, \cref{eq:kd-loss}. We discuss alternative ways to integrate outside information in \cref{sec:rel_work}.

\subsection{Combining neural topic modeling with knowledge distillation}\label{sec:methodology}

\paragraph{The knowledge distillation objective.} To apply KD to a ``base'' neural topic model, we replace the reconstruction term $\mathcal{L}_R$ in \cref{eq:elbo-recon}
with $\mathcal{L}_{KD}$, as follows:
\begin{align}
    \bm{w}_d^{\textsc{BAT}} &= \sigma\left( \bm{z}_d^{\textsc{BAT}} / T \right)N_d\nonumber\\
    \bm{\hat{w}} &= f(\bm{\theta}_d, \B; T)\nonumber \\
    \mathcal{L}_{KD} &= \lambda T^2\left(\bm{w}_d^{\textsc{BAT}}\right)^\top \log \bm{\hat{w}} + (1 - \lambda)\mathcal{L}_{R}\label{eq:kd-loss}
\end{align}
Here, $\bm{z}^{\textsc{BAT}}_d$ are the logits produced by the teacher network for a given input document $d$, meaning that $\bm{w}_d^{\textsc{BAT}}$ acts as a smoothed pseudo-document. $T$ is the softmax temperature, which controls how diffuse the estimated probability mass is over the words (hence $f(\cdot;T)$ is \cref{eq:word-prob} with the corresponding scaling). 
This differs from the original KD in two ways: (a) it scales the estimated probabilities by the document length $N_d$, and (b) it uses a multi-label loss. 

\paragraph{The teacher model.} We generate the teacher logits $\bm{z}^{\textsc{BAT}}$ using the pretrained transformer \textsc{DistilBERT} \citep{sanhDistilBERTDistilledVersion2019}, itself a distilled version of BERT \citep{devlin-etal-2019-bert}.\footnote{\textsc{DistilBERT}'s light weight accommodates longer documents, necessary for topic modeling. Even with this change, we divide very long documents into chunks, estimating logits for each chunk and taking the pointwise mean. More complex schemes \citep[i.e., LSTMs,][]{hochreiterLongShortTermMemory1997} yielded no benefit.}  BERT-like models are generally pretrained on large domain-general corpora with a language-modeling like objective, yielding an ability to capture nuances of linguistic context more effectively than bag-of-words models \citep{clark-etal-2019-bert,liu-etal-2019-linguistic,rogersPrimerBERTologyWhat2020}. Mirroring the NTM's formulation as a variational auto-encoder, we treat \textsc{DistilBERT} as a \emph{deterministic} auto-encoder, fine-tuning it with the document-reconstruction objective $\mathcal{L}_R$ on the same dataset. Thus, we use a \textbf{B}ERT-based \textbf{A}utoencoder as our \textbf{T}eacher model, hence \textbf{BAT}.\footnote{A reader familiar with variational NTMs may notice that we haven't mentioned an obvious means of incorporating representations from a pretrained transformer: encoding the document representation from a BERT-like model. This yields unimpressive results; see \cref{appendix:encoderbert}.}

\paragraph{Clipping the logit distribution.} Depending on preprocessing, $V$ may number in the tens of thousands of words. This leads to a long tail of probability mass assigned to unlikely terms, and breaks standard assumptions of sparsity. \citet{tangUnderstandingImprovingKnowledge2020}, working in a classification setting, find that truncating the logits to the top-$n$ classes and assigning uniform mass to the rest improves accuracy. We instead choose the top $c\,N_d, c\in\mathbb{R}^+$ logits and assign \emph{zero} probability to the remaining elements to enforce sparsity.

\section{Experimental Setup}\label{sec:experiment_setup}
\subsection{Data and Metrics}\label{sec:data}\label{sec:eval}

\begin{table}[]
  \footnotesize
  \centering
  \begin{tabular}{p{0.2in} p{0.2in} p{0.1in} p{0.4in} l}
    \toprule
                  & $D$   & $V$ & Avg $N_d$ & Preprocessing details                  \\
    \midrule
    \texttt{20NG} & 18k   & 2k  & 87.1      & \citet{Srivastava2017AutoencodingVI} \\
    \texttt{Wiki} & 28.5k & 20k & 1395.4    & \citet{nan-etal-2019-topic}          \\
    \texttt{IMDb} & 50k   & 5k  & 95.0      & \citet{card-etal-2018-neural}        \\
    \bottomrule
  \end{tabular}
  \caption{Corpus statistics, which vary in total number of documents ($D$), vocabulary size ($V$), and average document length ($N_d$).}
  \label{tab:data}
\end{table}

We validate our approach using three readily available datasets that vary widely in domain, corpus and vocabulary size, and document length: 20 Newsgroups \citep[\texttt{20NG}, ][]{langNewsWeederLearningFilter1995},\footnote{\url{ qwone.com/~jason/20Newsgroups}} Wikitext-103 \citep[\texttt{Wiki},][]{merityPointerSentinelMixture2017},\footnote{\url{ s3.amazonaws.com/research.metamind.io/wikitext/wikitext-103-v1.zip}} and IMDb movie reviews \citep[\texttt{IMDb},][]{maas-etal-2011-learning}.\footnote{\url{ ai.stanford.edu/~amaas/data/sentiment}} These are commonly used in neural topic modeling, with preprocessed versions provided by various authors; see references in \cref{tab:data} for details.
For consistency with prior work, we use a train/dev/test split of 48/12/40 for \texttt{20NG}, 70/15/15 for \texttt{Wiki}, and 50/25/25 for \texttt{IMDb}.\footnote{The splits are used to estimate NPMI. Dev splits are used to select hyperparameters, and test splits are run after hyperparameters are selected and frozen.}%

We seek to discover a latent space of topics that is meaningful and useful to people \citep{changReadingTeaLeaves2009}. Accordingly, we evaluate topic coherence using normalized mutual pointwise information (NPMI), which is significantly correlated with human judgments of topic quality \citep{aletras-stevenson-2013-evaluating,lau-etal-2014-machine} and widely used to evaluate topic models.\footnote{We also obtain competitive results for document perplexity, which has also been used widely but correlates negatively with human coherence evaluations \citep{changReadingTeaLeaves2009}.} We follow precedent and calculate (internal) NPMI using the top ten words in each topic, taking the mean across the NPMI scores for individual topics.  Internal NPMI is estimated with reference co-occurrence counts from a held-out dataset from the same corpus, i.e., the dev or test split. While internal NPMI is the metric of choice for most prior work, we also provide external NPMI results using Gigaword 5 \citep{parkerEnglishGigawordFifth2011a}, following \citet{card-etal-2018-neural}.\looseness=-1

\subsection{Experimental Baselines}\label{sec:baselines}

We select three experimental baseline models that represent diverse styles of neural topic modeling.\footnote{This use of ``baseline'' should not be confused with the ``base'' neural topic model augmented by knowledge distillation (\cref{sec:methodology}). }
Each achieves the highest NPMI on the majority of its respective datasets, as well as a considerable improvement over previous neural and non-neural topic models \citep[such as][]{Srivastava2017AutoencodingVI,miaoNeuralVariationalInference2016,ding-etal-2018-coherence}. All our baselines are roughly contemporaneous with one another, and had yet to be compared in a head-to-head fashion prior to our work.

\noindent {\bf \textsc{Scholar.}} \citet{card-etal-2018-neural} use a VAE-based \citep{kingmaAutoEncodingVariationalBayes2014} neural topic modeling setup \citep[as introduced in][]{Srivastava2017AutoencodingVI} with a logistic normal prior to approximate the Dirichlet, and provide an elegant way to incorporate document metadata.

\noindent {\bf \textsc{DVAE.}} \citet{burkhardtDecouplingSparsitySmoothness2019} use a Dirichlet prior, where its reparameterization is enabled by rejection sampling variational inference. This allows it to tap into the same generative story as the original LDA formulations of \citet{bleiLatentDirichletAllocation2003}, and to enjoy the advantageous properties of the Dirichlet like multi-modality \citep{wallachRethinkingLDAWhy2009,wangNeuralTopicModeling2020}.

\noindent {\bf \textsc{W-LDA.}} \citet{nan-etal-2019-topic}  forego the VAE in favor of a Wasserstein auto-encoder \citep{tolstikhinWassersteinAutoEncoders2018}, using a Dirichlet prior that is matched by minimizing Maximum Mean Discrepancy. They find the method leads to state-of-the-art coherence on several datasets and encourages topics to exhibit greater word diversity.

We demonstrate the modularity of our core innovation by combining our method with both \textsc{Scholar} and \textsc{W-LDA} (\cref{sec:results}).

\subsection{Our Models and Settings}\label{sec:our_model_settings}

As discussed in \cref{sec:methodology}, our approach relies on a ``base'' neural topic model and unnormalized probabilities over words estimated by a transformer as ``teacher''. We discuss each in turn.

\paragraph{Neural topic models augmented with knowledge distillation.} We experiment with both \textsc{Scholar} and \textsc{W-LDA} as base models. The former constitutes our primary model and point of comparison with baselines, while the latter is a proof-of-concept that attests to our method's modularity; we added knowledge distillation to \textsc{W-LDA} with only a few lines of code (\cref{appendix:code}). We evaluate both at $K=50$ and $K=200$ topics.

We tune using NPMI, with reference co-occurrence counts taken from a held-out development set from the relevant corpus. For our baselines, we use the publicly-released author implementations.\footnote{\textsc{Scholar}: \url{github.com/dallascard/scholar}\newline W-LDA: \url{github.com/awslabs/w-lda}\newline DVAE: \url{github.com/sophieburkhardt/dirichlet-vae-topic-models}\newline For augmented models we start with our own reimplementations of the baseline approaches in a common codebase, validated by obtaining comparable results to the original authors on their datasets. } While we generally attempt to retain the original hyperparameter settings when available, we do perform an exhaustive grid search on the \textsc{Scholar} baselines and \textsc{Scholar}+\ourmodel to ensure fairness in comparison (ranges, optimal values, and other details in \cref{appendix:hyperparam}).

Our method also introduces additional hyperparameters: the weight for KD loss, $\lambda$ (\cref{eq:kd-loss}); the softmax temperature $T$; and the proportion of the word-level teacher logits that we retain (relative to document length, see clipping in \cref{sec:methodology}). For most dataset-$K$ pairs, we find that we can improve topic quality under most settings, with a relatively small set of values for each hyperparameter leading to better results. In fact, following the extensive search on \textsc{Scholar}+BAT, we found we could tune \textsc{W-LDA} within a few iterations.

Topic models rely on random sampling procedures, and to ensure that our results are robust, we report the average values across five runs (previously unreported by the authors of our baselines).

\paragraph{The \textsc{DistilBERT} teacher.} We fine-tune a modified version of \textsc{DistilBERT} with the same document reconstruction objective as the NTM ($\mathcal{L}_R$, \cref{eq:elbo-recon}) on the training data.
Specifically, \textsc{DistilBERT} maps a WordPiece-tokenized \citep{wuGoogleNeuralMachine2016} document $d$ to an $l$-dimensional hidden vector with a transformer \citep{vaswaniAttentionAllYou2017a}, then back to logits over $V$ words (tokenized with the same scheme as the topic model).
For long documents, we split into blocks of 512 tokens and mean-pool the transformer outputs.
We use the pretrained model made available by the authors \citep{Wolf2019HuggingFacesTS}.
We train until perplexity converges on the same held-out dev set used in the topic modeling setting. Unsurprisingly, \textsc{DistilBERT} achieves dramatically lower perplexity than all topic model baselines. Note that we need only train the model once per corpus, and can experiment with different NTM variations using the same $\bm{z}^{\textsc{BAT}}$.

\begin{table*}[bhtp]
        \centering
        \resizebox{\textwidth}{!}{
                \begin{tabular}{@{}llllclll@{}}
                        \toprule
                        \multicolumn{1}{c}{\textbf{}} & \multicolumn{3}{c}{$K = 50$}      &                                   & \multicolumn{3}{c}{$K = 200$}                                                                                                                    \\ %
                        \multicolumn{1}{c}{}          & \multicolumn{1}{c}{\textbf{20NG}} & \multicolumn{1}{c}{\textbf{Wiki}} & \multicolumn{1}{c}{\textbf{IMDb}} &  & \multicolumn{1}{c}{\textbf{20NG}} & \multicolumn{1}{c}{\textbf{Wiki}} & \multicolumn{1}{c}{\textbf{IMDb}} \\ \midrule
                        \textbf{DVAE}                 & 0.340                             & 0.490                             & 0.145                             &  & 0.316                             & 0.450                             & 0.160                             \\
                        \textbf{W-LDA}                & 0.279                             & 0.494                             & 0.136                             &  & 0.188                             & 0.308                             & 0.095                             \\
                        \textbf{\textsc{Scholar}}     & 0.322 (0.007)                     & 0.494 (0.005)                     & 0.168 (0.002)                     &  & 0.263 (0.002)                     & 0.473 (0.005)                     & 0.140 (0.001)                     \\ \midrule
                        \textbf{\textsc{Sch. + BAT}}  & \textbf{0.354} (0.004)            & \textbf{0.521} (0.009)            & \textbf{0.182} (0.002)            &  & \textbf{0.332} (0.002)            & \textbf{0.513} (0.001)            & \textbf{0.175} (0.003)            \\ \bottomrule
                \end{tabular}
        }
        \caption{The NPMI for our baselines (\cref{sec:baselines}) compared with \ourmodel (explained in \cref{sec:methodology}) using \textsc{Scholar} as our base neural architecture. We achieve better NPMI than all baselines across three datasets and $K=50, K=200$ topics. %
                We use 5 random restarts and report the standard deviation.
        }
        \label{tab:results_main}
\end{table*}

\section{Results and Discussion}
\label{sec:results}

Using the VAE-based \textsc{Scholar} as the base model, topics discovered using \ourmodel are more coherent, as measured via NPMI, than previous state-of-the-art baseline NTMs (\cref{tab:results_main}), improving on the DVAE and W-LDA baselines, and the baseline of \textsc{Scholar} without the KD augmentation. 
We establish the robustness of our approach's improvement by taking the mean across multiple runs with different random seeds, yielding consistent improvement over all baselines for all the datasets. We validate the approach using a smaller and larger number of topics, $K = 50$ and $200$, respectively.

In addition to its improved performance, \ourmodel can apply straightforwardly to other models, because it makes very few assumptions about the base model---requiring only that it rely on a word-level reconstruction objective, which is true of the majority of neural topic models proposed to date. We illustrate this by using the Wasserstein auto-encoder (W-LDA) as a base NTM, showing in \cref{tab:results_wlda_kd} that \ourmodel improves on the unaugmented model.\footnote{We note that the W-LDA baseline did not tune well on 200 topics, further complicated by the model's extensive run time. As such, we focus on augmenting that model for 50 topics, consistent with the number of topics on which \citet{nan-etal-2019-topic} report their results. We add preliminary results using \ourmodel with DVAE in \cref{appendix:dvae}.} 

We report the dev set results (corresponding to the test set results in \cref{tab:results_main,tab:results_wlda_kd}) in \cref{appendix:test}---the same pattern of results is obtained, for all the models.\looseness=-1

Finally, we also compute NPMI using reference counts from an external corpus \citep[Gigaword 5,][]{parkerEnglishGigawordFifth2011a} for \textsc{Scholar} and \textsc{Scholar+BAT} (\cref{tab:results_external_npmi}). We find the same patterns generally hold: in all but one setting (\texttt{Wiki}, $K=50$), \ourmodel improves topic coherence relative to \textsc{Scholar}. These external NPMI results suggest that our model avails itself of the distilled general language knowledge from pretrained BERT, and moreover that our fine-tuning procedure does not overfit to the training data.\looseness=-1

\begin{table}[hbtp]
        \centering
        \resizebox{\columnwidth}{!}{
                \begin{tabular}{@{}rrrr@{}}
                        \toprule
                                       & \multicolumn{1}{c}{\textbf{20NG}} & \multicolumn{1}{c}{\textbf{Wiki}} & \multicolumn{1}{c}{\textbf{IMDb}} \\ \midrule
                        \textbf{W-LDA} & 0.279        (0.010)              & \textbf{0.494}     (0.012)        & 0.136      (0.008)                \\
                        \textbf{+BAT}  & \textbf{0.299}        (0.010)     & \textbf{0.505}   (0.014)          & \textbf{0.162}       (0.003)      \\\bottomrule
                \end{tabular}}
        \caption{
                Mean NPMI (s.d.) across 5 runs for W-LDA \cite{nan-etal-2019-topic} and W-LDA+\ourmodel for $K=50$, showing improvement on two of three datasets. This demonstrates that our method is \textit{modular} and can be used with base neural topic models that vary significantly in architecture.
        }
        \label{tab:results_wlda_kd}
\end{table}

\begin{table}[hbtp]
        \centering
        \begin{tabular}{r r l l }
                \toprule
                $K$   &               & \textbf{\textsc{Scholar}} & +\textbf{\ourmodel}    \\
                \midrule
                $50$  & \texttt{20ng} & 0.147 (0.002)             & \textbf{0.170} (0.006) \\
                      & \texttt{Wiki} & \textbf{0.193} (0.006)    & \textbf{0.187} (0.004) \\
                      & \texttt{IMDb} & 0.149 (0.003)             & \textbf{0.161} (0.003) \\

                $200$ & \texttt{20ng} & 0.111 (0.001)             & \textbf{0.171} (0.002) \\
                      & \texttt{Wiki} & 0.177 (0.003)             & \textbf{0.190} (0.008) \\
                      & \texttt{IMDb} & 0.122 (0.002)             & \textbf{0.159} (0.003) \\
                \midrule
        \end{tabular}
        \caption{External NPMI (s.d.) for the base \textsc{Scholar} and \textsc{Scholar}+BAT. Models selected according to performance on the development set using internal NPMI.
        }
        \label{tab:results_external_npmi}
\end{table}

\section{Impact of \ourmodel on Individual Topics}\label{sec:analysis}

Following standard practice, we have established that our models discover more coherent topics \emph{on average} when compared to others (\cref{tab:results_main,tab:results_wlda_kd}). Now, we look more closely at the extent to which those improvements are \emph{meaningful} at the level of individual topics. To do so we directly compare topics discovered by the baseline neural topic model (\textsc{Scholar}) with corresponding topics obtained when that model is augmented with \textsc{BAT}, looking at the NPMIs of the corresponding topics as well as considering them qualitatively.

We align the topics in the base and augmented \textsc{Scholar} models using a variation of competitive linking, which produces a greedy approximation to optimal weighted bipartite graph matching  \citep{melamed2000models}. A fully connected weighted bipartite graph is constructed by linking all topic pairs across (but not within) the two models, with the weight for a topic pair being the similarity between their word distributions as measured by Jenson-Shannon (JS) divergence \citep{wong1985entropy,lin1991divergence}. We pick the pair $(t_i,t_j)$ with the lowest JS divergence and add it to the resulting alignment, then remove $t_i$ and $t_j$ from consideration and iterate until no pairs are left. The resulting aligned topic pairs can then be sorted by their JS divergences to directly compare corresponding topics.\looseness=-1\footnote{Note that more similar topics have lower JS-divergence, so we are seeking to minimize rather than maximize total weight. We use JS-divergence because it is conveniently symmetric and finite.}

\begin{figure}[t!]
  \centering
  \includegraphics[scale=0.5]{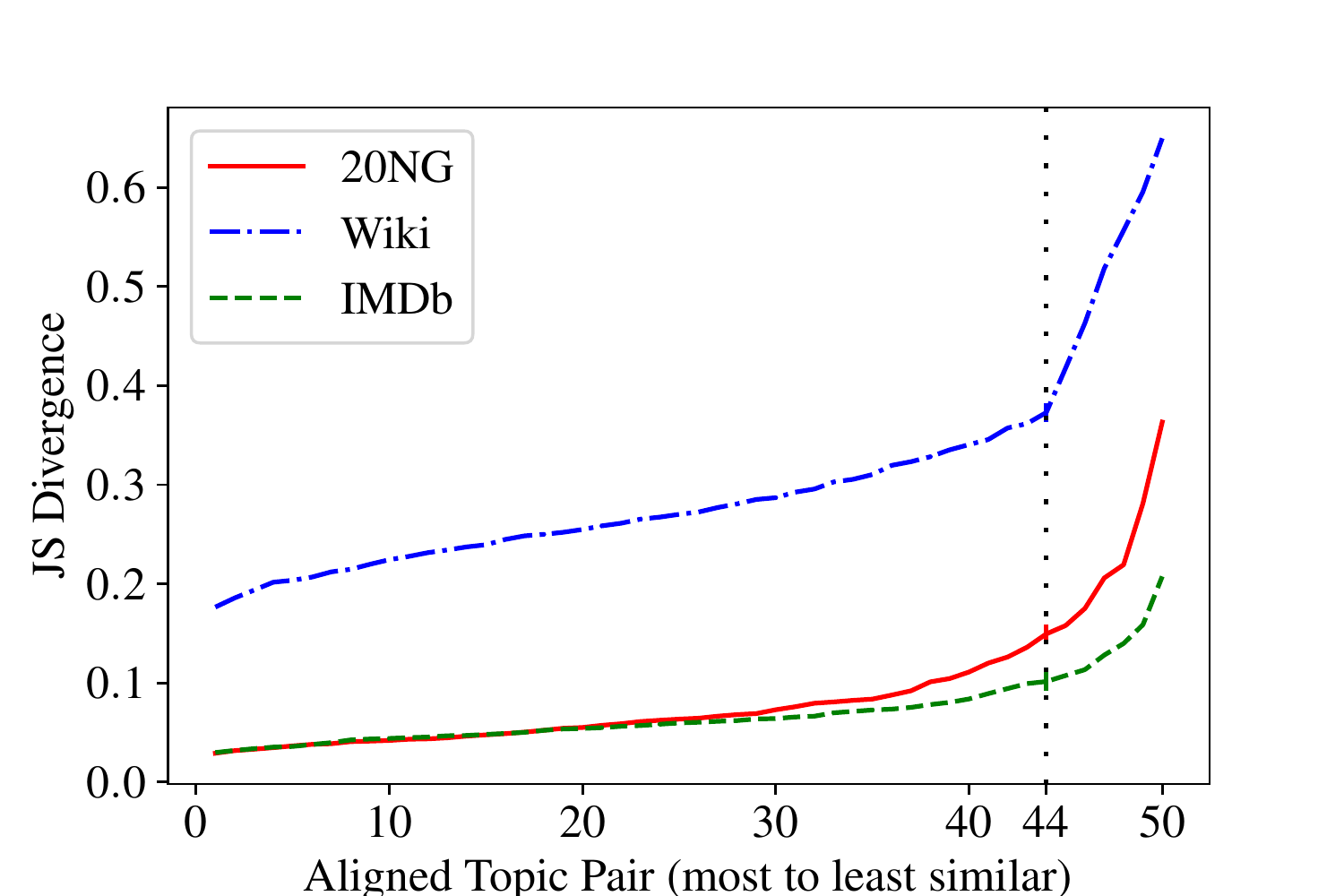}
  \caption{Jensen-Shannon divergence for aligned topic pairs in the \textsc{Scholar} and \textsc{Scholar}+\ourmodel models.
    }
  \label{fig:jsd}
\end{figure}

\cref{fig:jsd} shows the JS-divergences for aligned topic pairs, for our three corpora. Based on visual inspection, we choose the 44 most aligned topic pairs as being meaningful for comparison; beyond this point, the topics do not bear a conceptual relationship (using the same threshold for the three datasets for simplicity). %

\begin{figure}[t!]
  \centering
  \includegraphics[scale=0.5]{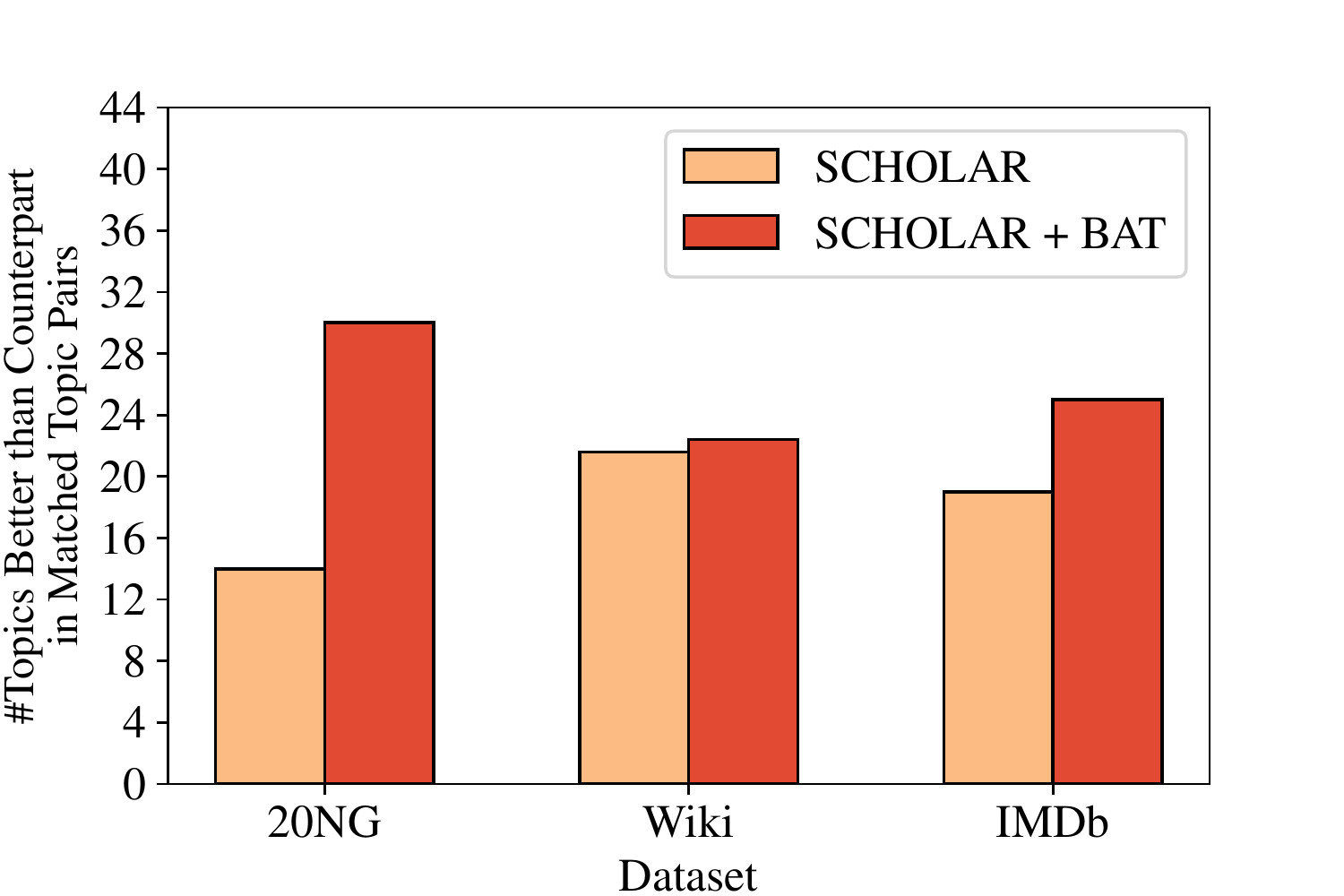}
  \caption{Number of matched topic pairs where \textsc{Scholar}+\ourmodel improves coherence, compared with the number of matched pairs where the baseline improves coherence.
    }
  \label{fig:bar_h2h_wins}
\end{figure}

When we consider these conceptually related topic pairs, we see that the model augmented with \ourmodel has the topic with the higher NPMI value more often across all three datasets (\cref{fig:bar_h2h_wins}). This means that \ourmodel is not just producing improvements in the aggregate (\cref{sec:results}): its effect can be interpreted more specifically as identifying the same space of topics generated by an existing model and, in most cases, improving the coherence of individual topics. This highlights the modular value of our approach.

\begin{table*}[tp]
  \centering
  \scalebox{0.8}{
    \begin{tabular}{ llrl }
      \toprule
                    &                                     & \textbf{NPMI} & \textbf{Topic}                                                                           \\ \midrule
      \texttt{20ng} & \textbf{\textsc{Scholar}}           & 0.454         & nhl hockey player coach ice playoff team league stanley european                         \\
                    & \textbf{\textsc{Scholar}+\ourmodel} & 0.523         & nhl hockey player team coach playoff cup wings stanley leafs                             \\ \midrule
      \texttt{Wiki} & \textbf{\textsc{Scholar}}           & 0.547         & jtwc jma typhoon monsoon luzon geophysical pagasa guam cyclone southwestward             \\
                    & \textbf{\textsc{Scholar}+\ourmodel} & 0.621         & jtwc jma typhoon meteorological intensification monsoon dissipating shear outflow trough \\ \midrule
      \texttt{IMDb} & \textbf{\textsc{Scholar}}           & 0.197         & adaptation version novel bbc versions jane kenneth handsome adaptations faithful         \\
                    & \textbf{\textsc{Scholar}+\ourmodel} & 0.218         & adaptation novel book read books faithful bbc version versions novels                    \\ \bottomrule
    \end{tabular}}
  \caption{Selected examples of \textsc{Scholar}+\ourmodel improving on topics from \textsc{Scholar}.
    We observe that the improved \texttt{20ng} topic is more cleanly focused on the NHL (removing \emph{european}, adding the Toronto Maple \emph{Leafs}, evoking the Stanley \emph{Cup} rather than the more generic \emph{ice}); the improved \texttt{wiki} topic about typhoons is more clearly concentrated on meterological terms, rather than interspersing specific locations of typhoons (\emph{luzon}, \emph{guam}); and the improved \texttt{IMDb} topic more cleanly reflects what we would characterize as ``video adaptations'' by bringing in terms about that subject (\emph{book}, \emph{books}, \emph{novels}, \emph{read}) in place of predominant words relating to particular adaptations.  Randomly selected examples can be found in \cref{appendix:analysis}.
  }
  \label{tab:topic_matched_ill_exs}
\end{table*}

\cref{tab:topic_matched_ill_exs} provides qualitative discussion for one example from each corpus, which we have selected for illustration from a single randomly selected run of the baseline \textsc{Scholar} and \textsc{Scholar}+\ourmodel models for $K=50$. We find that, consistent with prior work on automatic evaluation of topic models, differences in NPMI do appear to correspond to recognizable subjective differences in topic quality. So that readers may form their own judgments, \cref{appendix:analysis} presents 15 aligned pairs for each corpus, selected randomly by stratifying across levels of alignment quality to create a fair sample to review.

\section{Related Work}
\label{sec:rel_work}

\paragraph{Integrating embeddings into topic models.}
A key goal in our use of knowledge distillation is to incorporate relationships between words  that may not be well supported by the topic model's input documents alone. Some previous topic models have sought to address this issue by incorporating external word information, including word senses \cite{ferrugento-etal-2016-topic} and pretrained word embeddings \cite{hu2016latent,yang2017adapting,xunCorrelatedTopicModel2017,ding-etal-2018-coherence}.
More recently, \citet{bianchiPretrainingHotTopic2020} have incorporated BERT embeddings into the encoder to improve topic coherence. (See \cref{appendix:encoderbert} for our own related experiments, which yielded mixed results.) We refer the reader to \newcite{diengTopicModelingEmbedding2019} for an extensive and up-to-date overview.\looseness=-1

A limitation of these approaches is that they simply import general, non-corpus-specific word-level information. In contrast, representations from a pretrained transformer can benefit from both general language knowledge and corpus-dependent information, by way of the pretraining and fine-tuning regime. By regularizing toward representations conditioned on the document, we remain coherent relative to the topic model data.  An additional key advantage for our method is that it involves only a slight change to the underlying topic model, rather than the specialized designs by the above methods.

\paragraph{Knowledge distillation.} While the focus was originally on single-label image classification, KD has also been extended to the multi-label setting \cite{liuMultiLabelImageClassification2018}. In NLP, KD has usually been applied in supervised settings \cite{kim-rush-2016-sequence,huangKnowledgeDistillationSequence2018,yangTextBrewerOpenSourceKnowledge2020}, but also in some unsupervised tasks (usually using an unsupervised teacher for a supervised student) \cite{huCreatingSomethingNothing2020,sunKnowledgeDistillationMultilingual2020}. \citet{xuDistilledWassersteinLearning2018} use word embeddings jointly learned with a topic model in a procedure they term distillation, but do not follow the method from \citet{hintonDistillingKnowledgeNeural2015} that we employ (instead opting for joint-learning). Recently, pretrained models like BERT have offered an attractive choice of teacher model, used successfully for a variety of tasks such as sentiment classification and paraphrasing \cite{tangNaturalLanguageGeneration2019,tangDistillingTaskSpecificKnowledge2019}.
Work in distillation often cites a reduction in computational cost as a goal \cite[e.g.,][]{sanhDistilBERTDistilledVersion2019}, although we are aware of at least one effort that is focused specifically on interpretability \cite{liuImprovingInterpretabilityDeep2018}.\looseness=-1

\paragraph{Topic diversity.}
Coherence, commonly quantified automatically using NPMI, is the current standard for evaluating topic model quality. Recently several authors \citep{diengTopicModelingEmbedding2019,burkhardtDecouplingSparsitySmoothness2019,nan-etal-2019-topic} have proposed additional metrics focused on the diversity or uniqueness of topics (based on top words in topics). However, no one metric has yet achieved acceptance or consensus in the literature. Moreover, such measures fail to distinguish between the case where two topics share the same set of top $n$ words, therefore coming across as essentially identical, versus when one topic's top $n$ words are repeated individually across multiple other topics, indicating a weaker and more diffuse similarity to those topics. We discuss issues related to topic diversity in \cref{appendix:tu}.

\section{Conclusions and Future Work}\label{sec:conclusion}

To our knowledge, we are the first to distill a ``black-box'' neural network teacher to guide a probabilistic graphical model. We do this in order to combine the expressivity of probabilistic topic models with the precision of pretrained transformers. Our modular method sits atop any neural topic model (NTM) to improve topic quality, which we demonstrate using two NTMs of highly disparate architectures (VAEs and WAEs), obtaining state-of-the-art topic coherence across three datasets from different domains. Our adaptable framework does not just produce improvements in the aggregate (as is commonly reported): its effect can be interpreted more specifically as identifying the same space of topics generated by an existing model and, in most cases, improving the coherence of individual topics, thus highlighting the modular value of our approach.

In future work, we also hope to explore the effects of the pretraining corpus \cite{gururangan-etal-2020-dont} and teachers (besides BERT) on the generated topics.
Another intriguing direction is exploring the connection between our methods and neural network interpretability. The use of knowledge distillation to facilitate interpretability has also been previously explored, for example, in \newcite{liuImprovingInterpretabilityDeep2018} to learn interpretable decision trees from neural networks. In our work, as the weight on the BERT autoencoder logits $\lambda$ goes to one, the topic model begins to describe less the \emph{corpus} and more the \emph{teacher}. We believe mining this connection can open up further research avenues; for instance, by investigating the differences in such teacher-topics conditioned on the pre-training corpus. Finally, although we are motivated primarily by the widespread use of topic models for identifying interpretable topics \cite[][Ch. 3]{boyd2017applications},  we plan to explore the ideas presented here further in the context of downstream applications like document classification.  \looseness=-1

\section*{Acknowledgements}
This material is based upon work supported by the National Science Foundation under Grants 2031736 and 2008761. We thank Pedro Rodriguez, Shi Feng, and our anonymous reviewers for their helpful comments. Appreciation to Adam Forbes for the design of \cref{fig:thresh_matched_topics}.  We also thank the authors of \citet{card-etal-2018-neural}, \citet{nan-etal-2019-topic}, and \citet{burkhardtDecouplingSparsitySmoothness2019} for their publicly available implementations.

\bibliography{anthology,refs,zotero}

\begin{thebibliography}{58}
\expandafter\ifx\csname natexlab\endcsname\relax\def\natexlab#1{#1}\fi

\bibitem[{Aletras and Stevenson(2013)}]{aletras-stevenson-2013-evaluating}
Nikolaos Aletras and Mark Stevenson. 2013.
\newblock \href {https://www.aclweb.org/anthology/W13-0102} {Evaluating topic
  coherence using distributional semantics}.
\newblock In \emph{Proceedings of the 10th International Conference on
  Computational Semantics ({IWCS} 2013) {--} Long Papers}, pages 13--22,
  Potsdam, Germany. Association for Computational Linguistics.

\bibitem[{Bianchi et~al.(2020)Bianchi, Terragni, and
  Hovy}]{bianchiPretrainingHotTopic2020}
Federico Bianchi, Silvia Terragni, and Dirk Hovy. 2020.
\newblock \href {http://arxiv.org/abs/2004.03974} {Pre-training is a {{Hot
  Topic}}: {{Contextualized Document Embeddings Improve Topic Coherence}}}.
\newblock \emph{arXiv:2004.03974 [cs]}.

\bibitem[{Blei(2003)}]{bleiLatentDirichletAllocation2003}
David~M Blei. 2003.
\newblock Latent {{Dirichlet Allocation}}.
\newblock \emph{Journal of Machine Learning Research}, page~30.

\bibitem[{Boyd-Graber et~al.(2017)Boyd-Graber, Hu, and
  Mimno}]{boyd2017applications}
Jordan Boyd-Graber, Yuening Hu, and David Mimno. 2017.
\newblock \href {http://www.nowpublishers.com/article/Details/INR-030}
  {\emph{Applications of Topic Models}}, volume~11 of \emph{Foundations and
  Trends in Information Retrieval}.
\newblock NOW Publishers.

\bibitem[{Burkhardt and
  Kramer(2019)}]{burkhardtDecouplingSparsitySmoothness2019}
Sophie Burkhardt and Stefan Kramer. 2019.
\newblock Decoupling {{Sparsity}} and {{Smoothness}} in the {{Dirichlet
  Variational Autoencoder Topic Model}}.
\newblock \emph{Journal of Machine Learning Research}, 20(131):27.

\bibitem[{Card et~al.(2018)Card, Tan, and Smith}]{card-etal-2018-neural}
Dallas Card, Chenhao Tan, and Noah~A. Smith. 2018.
\newblock \href {https://doi.org/10.18653/v1/P18-1189} {Neural models for
  documents with metadata}.
\newblock In \emph{Proceedings of the 56th Annual Meeting of the Association
  for Computational Linguistics (Volume 1: Long Papers)}, pages 2031--2040,
  Melbourne, Australia. Association for Computational Linguistics.

\bibitem[{Chang et~al.(2009)Chang, Gerrish, Wang, {Boyd-graber}, and
  Blei}]{changReadingTeaLeaves2009}
Jonathan Chang, Sean Gerrish, Chong Wang, Jordan~L. {Boyd-graber}, and David~M.
  Blei. 2009.
\newblock Reading {{Tea Leaves}}: {{How Humans Interpret Topic Models}}.
\newblock In \emph{Advances in {{Neural Information Processing Systems}} 22},
  pages 288--296. {Curran Associates, Inc.}

\bibitem[{Clark et~al.(2019)Clark, Khandelwal, Levy, and
  Manning}]{clark-etal-2019-bert}
Kevin Clark, Urvashi Khandelwal, Omer Levy, and Christopher~D. Manning. 2019.
\newblock \href {https://doi.org/10.18653/v1/W19-4828} {What does {BERT} look
  at? an analysis of {BERT}{'}s attention}.
\newblock In \emph{Proceedings of the 2019 ACL Workshop BlackboxNLP: Analyzing
  and Interpreting Neural Networks for NLP}, pages 276--286, Florence, Italy.
  Association for Computational Linguistics.

\bibitem[{Devlin et~al.(2019)Devlin, Chang, Lee, and
  Toutanova}]{devlin-etal-2019-bert}
Jacob Devlin, Ming-Wei Chang, Kenton Lee, and Kristina Toutanova. 2019.
\newblock \href {https://doi.org/10.18653/v1/N19-1423} {{BERT}: Pre-training of
  deep bidirectional transformers for language understanding}.
\newblock In \emph{Proceedings of the 2019 Conference of the North {A}merican
  Chapter of the Association for Computational Linguistics: Human Language
  Technologies, Volume 1 (Long and Short Papers)}, pages 4171--4186,
  Minneapolis, Minnesota. Association for Computational Linguistics.

\bibitem[{Dieng et~al.(2020)Dieng, Ruiz, and
  Blei}]{diengTopicModelingEmbedding2019}
Adji~B Dieng, Francisco~JR Ruiz, and David~M Blei. 2020.
\newblock Topic {{Modeling}} in {{Embedding Spaces}}.
\newblock \emph{Transactions of the Association for Computational Linguistics},
  8:439--453.

\bibitem[{Ding et~al.(2018)Ding, Nallapati, and
  Xiang}]{ding-etal-2018-coherence}
Ran Ding, Ramesh Nallapati, and Bing Xiang. 2018.
\newblock \href {https://doi.org/10.18653/v1/D18-1096} {Coherence-aware neural
  topic modeling}.
\newblock In \emph{Proceedings of the 2018 Conference on Empirical Methods in
  Natural Language Processing}, pages 830--836, Brussels, Belgium. Association
  for Computational Linguistics.

\bibitem[{Ferrugento et~al.(2016)Ferrugento, Oliveira, Alves, and
  Rodrigues}]{ferrugento-etal-2016-topic}
Adriana Ferrugento, Hugo~Gon{\c{c}}alo Oliveira, Ana Alves, and Filipe
  Rodrigues. 2016.
\newblock \href {https://www.aclweb.org/anthology/L16-1540} {Can topic
  modelling benefit from word sense information?}
\newblock In \emph{Proceedings of the Tenth International Conference on
  Language Resources and Evaluation ({LREC} 2016)}, pages 3387--3393,
  Portoro{\v{z}}, Slovenia. European Language Resources Association (ELRA).

\bibitem[{Figurnov et~al.(2018)Figurnov, Mohamed, and
  Mnih}]{Figurnov2018ImplicitRG}
Mikhail Figurnov, S.~Mohamed, and A.~Mnih. 2018.
\newblock Implicit reparameterization gradients.
\newblock In \emph{NeurIPS}.

\bibitem[{Gururangan et~al.(2020)Gururangan, Marasovi{\'c}, Swayamdipta, Lo,
  Beltagy, Downey, and Smith}]{gururangan-etal-2020-dont}
Suchin Gururangan, Ana Marasovi{\'c}, Swabha Swayamdipta, Kyle Lo, Iz~Beltagy,
  Doug Downey, and Noah~A. Smith. 2020.
\newblock \href {https://doi.org/10.18653/v1/2020.acl-main.740} {Don{'}t stop
  pretraining: Adapt language models to domains and tasks}.
\newblock In \emph{Proceedings of the 58th Annual Meeting of the Association
  for Computational Linguistics}, pages 8342--8360, Online. Association for
  Computational Linguistics.

\bibitem[{Hinton et~al.(2015)Hinton, Vinyals, and
  Dean}]{hintonDistillingKnowledgeNeural2015}
Geoffrey Hinton, Oriol Vinyals, and Jeffrey Dean. 2015.
\newblock Distilling the knowledge in a neural network.
\newblock In \emph{{{NIPS}} Deep Learning and Representation Learning
  Workshop}.

\bibitem[{Hochreiter and Schmidhuber(1997)}]{hochreiterLongShortTermMemory1997}
Sepp Hochreiter and J{\"u}rgen Schmidhuber. 1997.
\newblock \href {https://doi.org/10.1162/neco.1997.9.8.1735} {Long
  {{Short}}-{{Term Memory}}}.
\newblock \emph{Neural Computation}, 9(8):1735--1780.

\bibitem[{Hu et~al.(2020)Hu, Xie, Hong, and
  Tian}]{huCreatingSomethingNothing2020}
Hengtong Hu, Lingxi Xie, Richang Hong, and Qi~Tian. 2020.
\newblock Creating {{Something}} from {{Nothing}}: {{Unsupervised Knowledge
  Distillation}} for {{Cross}}-{{Modal Hashing}}.
\newblock In \emph{Proceedings of the IEEE/CVF Conference on Computer Vision
  and Pattern Recognition}, pages 3123--3132.

\bibitem[{Hu and Tsujii(2016)}]{hu2016latent}
Weihua Hu and Jun’ichi Tsujii. 2016.
\newblock A latent concept topic model for robust topic inference using word
  embeddings.
\newblock In \emph{Proceedings of the 54th Annual Meeting of the Association
  for Computational Linguistics (Volume 2: Short Papers)}, pages 380--386.

\bibitem[{Huang et~al.(2018)Huang, You, Chen, Qian, and
  Yu}]{huangKnowledgeDistillationSequence2018}
Mingkun Huang, Yongbin You, Zhehuai Chen, Yanmin Qian, and Kai Yu. 2018.
\newblock \href {https://doi.org/10.21437/Interspeech.2018-1589} {Knowledge
  {{Distillation}} for {{Sequence Model}}}.
\newblock In \emph{Interspeech 2018}, pages 3703--3707. {ISCA}.

\bibitem[{Jankowiak and
  Obermeyer(2018)}]{jankowiakPathwiseDerivativesReparameterization2018}
Martin Jankowiak and Fritz Obermeyer. 2018.
\newblock Pathwise {{Derivatives Beyond}} the {{Reparameterization Trick}}.
\newblock In \emph{International Conference on Machine Learning}, pages
  2235--2244.

\bibitem[{Kim and Rush(2016)}]{kim-rush-2016-sequence}
Yoon Kim and Alexander~M. Rush. 2016.
\newblock \href {https://doi.org/10.18653/v1/D16-1139} {Sequence-level
  knowledge distillation}.
\newblock In \emph{Proceedings of the 2016 Conference on Empirical Methods in
  Natural Language Processing}, pages 1317--1327, Austin, Texas. Association
  for Computational Linguistics.

\bibitem[{Kingma and Welling(2014)}]{kingmaAutoEncodingVariationalBayes2014}
Diederik~P. Kingma and Max Welling. 2014.
\newblock \href {http://arxiv.org/abs/1312.6114} {Auto-{{Encoding Variational
  Bayes}}}.
\newblock In \emph{2nd International Conference on Learning Representations,
  {ICLR} 2014, Banff, AB, Canada, April 14-16, 2014, Conference Track
  Proceedings}.

\bibitem[{Landauer and Dumais(1997)}]{landauerSolutionPlatoProblem1997}
Thomas~K. Landauer and Susan~T. Dumais. 1997.
\newblock \href {https://doi.org/10.1037/0033-295X.104.2.211} {A solution to
  {{Plato}}'s problem: {{The}} latent semantic analysis theory of acquisition,
  induction, and representation of knowledge}.
\newblock \emph{Psychological Review}, 104(2):211--240.

\bibitem[{Lang(1995)}]{langNewsWeederLearningFilter1995}
Ken Lang. 1995.
\newblock \href {https://doi.org/10.1016/B978-1-55860-377-6.50048-7}
  {{{NewsWeeder}}: {{Learning}} to {{Filter Netnews}}}.
\newblock In Armand Prieditis and Stuart Russell, editors, \emph{Machine
  {{Learning Proceedings}} 1995}, pages 331--339. {Morgan Kaufmann}, {San
  Francisco (CA)}.

\bibitem[{Lau et~al.(2014)Lau, Newman, and Baldwin}]{lau-etal-2014-machine}
Jey~Han Lau, David Newman, and Timothy Baldwin. 2014.
\newblock \href {https://doi.org/10.3115/v1/E14-1056} {Machine reading tea
  leaves: Automatically evaluating topic coherence and topic model quality}.
\newblock In \emph{Proceedings of the 14th Conference of the {E}uropean Chapter
  of the Association for Computational Linguistics}, pages 530--539,
  Gothenburg, Sweden. Association for Computational Linguistics.

\bibitem[{Lin(1991)}]{lin1991divergence}
Jianhua Lin. 1991.
\newblock Divergence measures based on the shannon entropy.
\newblock \emph{IEEE Transactions on Information theory}, 37(1):145--151.

\bibitem[{Liu et~al.(2019)Liu, Gardner, Belinkov, Peters, and
  Smith}]{liu-etal-2019-linguistic}
Nelson~F. Liu, Matt Gardner, Yonatan Belinkov, Matthew~E. Peters, and Noah~A.
  Smith. 2019.
\newblock \href {https://doi.org/10.18653/v1/N19-1112} {Linguistic knowledge
  and transferability of contextual representations}.
\newblock In \emph{Proceedings of the 2019 Conference of the North {A}merican
  Chapter of the Association for Computational Linguistics: Human Language
  Technologies, Volume 1 (Long and Short Papers)}, pages 1073--1094,
  Minneapolis, Minnesota. Association for Computational Linguistics.

\bibitem[{Liu et~al.(2018{\natexlab{a}})Liu, Wang, and
  Matwin}]{liuImprovingInterpretabilityDeep2018}
Xuan Liu, Xiaoguang Wang, and Stan Matwin. 2018{\natexlab{a}}.
\newblock Improving the {{Interpretability}} of {{Deep Neural Networks}} with
  {{Knowledge Distillation}}.
\newblock In \emph{2018 IEEE International Conference on Data Mining Workshops
  (ICDMW)}, pages 905--912. IEEE.

\bibitem[{Liu et~al.(2018{\natexlab{b}})Liu, Sheng, Shao, Yan, Xiang, and
  Pan}]{liuMultiLabelImageClassification2018}
Yongcheng Liu, Lu~Sheng, Jing Shao, Junjie Yan, Shiming Xiang, and Chunhong
  Pan. 2018{\natexlab{b}}.
\newblock \href {https://doi.org/10.1145/3240508.3240567} {Multi-{{Label Image
  Classification}} via {{Knowledge Distillation}} from {{Weakly}}-{{Supervised
  Detection}}}.
\newblock \emph{2018 ACM Multimedia Conference on Multimedia Conference - MM
  '18}, pages 700--708.

\bibitem[{Maas et~al.(2011)Maas, Daly, Pham, Huang, Ng, and
  Potts}]{maas-etal-2011-learning}
Andrew~L. Maas, Raymond~E. Daly, Peter~T. Pham, Dan Huang, Andrew~Y. Ng, and
  Christopher Potts. 2011.
\newblock \href {https://www.aclweb.org/anthology/P11-1015} {Learning word
  vectors for sentiment analysis}.
\newblock In \emph{Proceedings of the 49th Annual Meeting of the Association
  for Computational Linguistics: Human Language Technologies}, pages 142--150,
  Portland, Oregon, USA. Association for Computational Linguistics.

\bibitem[{Melamed(2000)}]{melamed2000models}
I~Dan Melamed. 2000.
\newblock Models of translational equivalence among words.
\newblock \emph{Computational Linguistics}, 26(2):221--249.

\bibitem[{Merity et~al.(2017)Merity, Xiong, Bradbury, and
  Socher}]{merityPointerSentinelMixture2017}
Stephen Merity, Caiming Xiong, James Bradbury, and Richard Socher. 2017.
\newblock Pointer {{Sentinel Mixture Models}}.
\newblock \emph{ICLR}.

\bibitem[{Miao et~al.(2016)Miao, Yu, and
  Blunsom}]{miaoNeuralVariationalInference2016}
Yishu Miao, Lei Yu, and Phil Blunsom. 2016.
\newblock \href {http://arxiv.org/abs/1511.06038} {Neural {{Variational
  Inference}} for {{Text Processing}}}.
\newblock \emph{ICML}.

\bibitem[{Mikolov et~al.(2013)Mikolov, Chen, Corrado, and
  Dean}]{mikolovEfficientEstimationWord2013}
Tomas Mikolov, Kai Chen, Greg Corrado, and Jeffrey Dean. 2013.
\newblock \href {http://arxiv.org/abs/1301.3781} {Efficient {{Estimation}} of
  {{Word Representations}} in {{Vector Space}}}.
\newblock In \emph{1st International Conference on Learning Representations,
  {ICLR} 2013, Scottsdale, Arizona, USA, May 2-4, 2013, Workshop Track
  Proceedings}.

\bibitem[{Nan et~al.(2019)Nan, Ding, Nallapati, and
  Xiang}]{nan-etal-2019-topic}
Feng Nan, Ran Ding, Ramesh Nallapati, and Bing Xiang. 2019.
\newblock \href {https://doi.org/10.18653/v1/P19-1640} {Topic modeling with
  {W}asserstein autoencoders}.
\newblock In \emph{Proceedings of the 57th Annual Meeting of the Association
  for Computational Linguistics}, pages 6345--6381, Florence, Italy.
  Association for Computational Linguistics.

\bibitem[{Nguyen et~al.(2013)Nguyen, Ying, and Resnik}]{nguyen2013lexical}
Viet-An Nguyen, Jordan~L Ying, and Philip Resnik. 2013.
\newblock Lexical and hierarchical topic regression.
\newblock In \emph{Advances in neural information processing systems}, pages
  1106--1114.

\bibitem[{Parker et~al.(2011)Parker, Graff, Kong, Chen, and
  Maeda}]{parkerEnglishGigawordFifth2011a}
Robert Parker, David Graff, Junbo Kong, Ke~Chen, and Kazuaki Maeda. 2011.
\newblock English {{Gigaword Fifth Edition}}.

\bibitem[{Pedregosa et~al.(2011)Pedregosa, Varoquaux, Gramfort, Michel,
  Thirion, Grisel, Blondel, Prettenhofer, Weiss, Dubourg, Vanderplas, Passos,
  Cournapeau, Brucher, Perrot, and Duchesnay}]{scikit-learn}
F.~Pedregosa, G.~Varoquaux, A.~Gramfort, V.~Michel, B.~Thirion, O.~Grisel,
  M.~Blondel, P.~Prettenhofer, R.~Weiss, V.~Dubourg, J.~Vanderplas, A.~Passos,
  D.~Cournapeau, M.~Brucher, M.~Perrot, and E.~Duchesnay. 2011.
\newblock Scikit-learn: Machine learning in {P}ython.
\newblock \emph{Journal of Machine Learning Research}, 12:2825--2830.

\bibitem[{Raffel et~al.(2019)Raffel, Shazeer, Roberts, Lee, Narang, Matena,
  Zhou, Li, and Liu}]{raffelExploringLimitsTransfer2019}
Colin Raffel, Noam Shazeer, Adam Roberts, Katherine Lee, Sharan Narang, Michael
  Matena, Yanqi Zhou, Wei Li, and Peter~J. Liu. 2019.
\newblock \href {https://arxiv.org/pdf/1910.10683.pdf} {Exploring the
  {{Limits}} of {{Transfer Learning}} with a {{Unified Text}}-to-{{Text
  Transformer}}}.
\newblock \emph{Journal of Machine Learning Research}.

\bibitem[{Rogers et~al.(2020)Rogers, Kovaleva, and
  Rumshisky}]{rogersPrimerBERTologyWhat2020}
Anna Rogers, Olga Kovaleva, and Anna Rumshisky. 2020.
\newblock \href {http://arxiv.org/abs/2002.12327} {A {{Primer}} in
  {{BERTology}}: {{What}} we know about how {{BERT}} works}.
\newblock \emph{arXiv:2002.12327 [cs]}.

\bibitem[{Sanh et~al.(2019)Sanh, Debut, Chaumond, and
  Wolf}]{sanhDistilBERTDistilledVersion2019}
Victor Sanh, Lysandre Debut, Julien Chaumond, and Thomas Wolf. 2019.
\newblock \href {http://arxiv.org/abs/1910.01108} {{{DistilBERT}}, a distilled
  version of {{BERT}}: Smaller, faster, cheaper and lighter}.
\newblock \emph{NeurIPS EMC\^2 Workshop}.

\bibitem[{Srivastava and Sutton(2017)}]{Srivastava2017AutoencodingVI}
Akash Srivastava and Charles Sutton. 2017.
\newblock Autoencoding variational inference for topic models.
\newblock In \emph{{{ICLR}}}.

\bibitem[{Sun et~al.(2020)Sun, Wang, Chen, Utiyama, Sumita, and
  Zhao}]{sunKnowledgeDistillationMultilingual2020}
Haipeng Sun, Rui Wang, Kehai Chen, Masao Utiyama, Eiichiro Sumita, and Tiejun
  Zhao. 2020.
\newblock \href {http://arxiv.org/abs/2004.10171} {Knowledge {{Distillation}}
  for {{Multilingual Unsupervised Neural Machine Translation}}}.
\newblock \emph{arXiv:2004.10171 [cs]}.

\bibitem[{Tang et~al.(2020)Tang, Shivanna, Zhao, Lin, Singh, Chi, and
  Jain}]{tangUnderstandingImprovingKnowledge2020}
Jiaxi Tang, Rakesh Shivanna, Zhe Zhao, Dong Lin, Anima Singh, Ed~H. Chi, and
  Sagar Jain. 2020.
\newblock \href {http://arxiv.org/abs/2002.03532} {Understanding and
  {{Improving Knowledge Distillation}}}.
\newblock \emph{arXiv:2002.03532 [cs, stat]}.

\bibitem[{Tang et~al.(2019{\natexlab{a}})Tang, Lu, and
  Lin}]{tangNaturalLanguageGeneration2019}
Raphael Tang, Yao Lu, and Jimmy Lin. 2019{\natexlab{a}}.
\newblock \href {https://doi.org/10.18653/v1/D19-6122} {Natural {{Language
  Generation}} for {{Effective Knowledge Distillation}}}.
\newblock In \emph{Proceedings of the 2nd {{Workshop}} on {{Deep Learning
  Approaches}} for {{Low}}-{{Resource NLP}} ({{DeepLo}} 2019)}, pages 202--208,
  {Hong Kong, China}. {Association for Computational Linguistics}.

\bibitem[{Tang et~al.(2019{\natexlab{b}})Tang, Lu, Liu, Mou, Vechtomova, and
  Lin}]{tangDistillingTaskSpecificKnowledge2019}
Raphael Tang, Yao Lu, Linqing Liu, Lili Mou, Olga Vechtomova, and Jimmy Lin.
  2019{\natexlab{b}}.
\newblock \href {http://arxiv.org/abs/1903.12136} {Distilling
  {{Task}}-{{Specific Knowledge}} from {{BERT}} into {{Simple Neural
  Networks}}}.
\newblock \emph{arXiv:1903.12136 [cs]}.

\bibitem[{Tolstikhin et~al.(2018)Tolstikhin, Gelly, Bousquet, and
  Scholkopf}]{tolstikhinWassersteinAutoEncoders2018}
Ilya Tolstikhin, Sylvain Gelly, Olivier Bousquet, and Bernhard Scholkopf. 2018.
\newblock Wasserstein {{Auto}}-{{Encoders}}.
\newblock \emph{ICLR}, page~16.

\bibitem[{Vaswani et~al.(2017)Vaswani, Shazeer, Parmar, Uszkoreit, Jones,
  Gomez, Kaiser, and Polosukhin}]{vaswaniAttentionAllYou2017a}
Ashish Vaswani, Noam Shazeer, Niki Parmar, Jakob Uszkoreit, Llion Jones,
  Aidan~N Gomez, \L~ukasz Kaiser, and Illia Polosukhin. 2017.
\newblock Attention is {{All}} you {{Need}}.
\newblock In I.~Guyon, U.~V. Luxburg, S.~Bengio, H.~Wallach, R.~Fergus,
  S.~Vishwanathan, and R.~Garnett, editors, \emph{Advances in {{Neural
  Information Processing Systems}} 30}, pages 5998--6008. {Curran Associates,
  Inc.}

\bibitem[{Wallach et~al.(2009)Wallach, Mimno, and
  McCallum}]{wallachRethinkingLDAWhy2009}
Hanna~M. Wallach, David~M. Mimno, and Andrew McCallum. 2009.
\newblock Rethinking {{LDA}}: {{Why Priors Matter}}.
\newblock In Y.~Bengio, D.~Schuurmans, J.~D. Lafferty, C.~K.~I. Williams, and
  A.~Culotta, editors, \emph{Advances in {{Neural Information Processing
  Systems}} 22}, pages 1973--1981. {Curran Associates, Inc.}

\bibitem[{Wang et~al.(2020)Wang, Hu, Zhou, He, Xiong, Ye, and
  Xu}]{wangNeuralTopicModeling2020}
Rui Wang, Xuemeng Hu, Deyu Zhou, Yulan He, Yuxuan Xiong, Chenchen Ye, and
  Haiyang Xu. 2020.
\newblock Neural {{Topic Modeling}} with {{Bidirectional Adversarial
  Training}}.
\newblock \emph{ACL}, page~11.

\bibitem[{Wang et~al.(2019)Wang, Zhou, and
  He}]{wangATMAdversarialneuralTopic2019}
Rui Wang, Deyu Zhou, and Yulan He. 2019.
\newblock {{ATM}}:{{Adversarial}}-neural {{Topic Model}}.
\newblock \emph{Information Processing \& Management}, 56(6):102098.

\bibitem[{Wolf et~al.(2019)Wolf, Debut, Sanh, Chaumond, Delangue, Moi, Cistac,
  Rault, Louf, Funtowicz, and Brew}]{Wolf2019HuggingFacesTS}
Thomas Wolf, Lysandre Debut, Victor Sanh, Julien Chaumond, Clement Delangue,
  Anthony Moi, Pierric Cistac, Tim Rault, R'emi Louf, Morgan Funtowicz, and
  Jamie Brew. 2019.
\newblock {{HuggingFace}}'s transformers: {{State}}-of-the-art natural language
  processing.
\newblock \emph{ArXiv}, abs/1910.03771.

\bibitem[{Wong and You(1985)}]{wong1985entropy}
Andrew~KC Wong and Manlai You. 1985.
\newblock Entropy and distance of random graphs with application to structural
  pattern recognition.
\newblock \emph{IEEE Transactions on Pattern Analysis and Machine
  Intelligence}, (5):599--609.

\bibitem[{Wu et~al.(2016)Wu, Schuster, Chen, Le, Norouzi, Macherey, Krikun,
  Cao, Gao, Macherey, Klingner, Shah, Johnson, Liu, Kaiser, Gouws, Kato, Kudo,
  Kazawa, Stevens, Kurian, Patil, Wang, Young, Smith, Riesa, Rudnick, Vinyals,
  Corrado, Hughes, and Dean}]{wuGoogleNeuralMachine2016}
Yonghui Wu, Mike Schuster, Zhifeng Chen, Quoc~V. Le, Mohammad Norouzi, Wolfgang
  Macherey, Maxim Krikun, Yuan Cao, Qin Gao, Klaus Macherey, Jeff Klingner,
  Apurva Shah, Melvin Johnson, Xiaobing Liu, \L~ukasz Kaiser, Stephan Gouws,
  Yoshikiyo Kato, Taku Kudo, Hideto Kazawa, Keith Stevens, George Kurian,
  Nishant Patil, Wei Wang, Cliff Young, Jason Smith, Jason Riesa, Alex Rudnick,
  Oriol Vinyals, Greg Corrado, Macduff Hughes, and Jeffrey Dean. 2016.
\newblock \href {http://arxiv.org/abs/1609.08144} {Google's {{Neural Machine
  Translation System}}: {{Bridging}} the {{Gap}} between {{Human}} and
  {{Machine Translation}}}.
\newblock \emph{arXiv:1609.08144 [cs]}.

\bibitem[{Xu et~al.(2018)Xu, Wang, Liu, and
  Carin}]{xuDistilledWassersteinLearning2018}
Hongteng Xu, Wenlin Wang, Wei Liu, and Lawrence Carin. 2018.
\newblock Distilled {{Wasserstein Learning}} for {{Word Embedding}} and {{Topic
  Modeling}}.
\newblock In S.~Bengio, H.~Wallach, H.~Larochelle, K.~Grauman,
  N.~{Cesa-Bianchi}, and R.~Garnett, editors, \emph{Advances in {{Neural
  Information Processing Systems}} 31}, pages 1716--1725. {Curran Associates,
  Inc.}

\bibitem[{Xun et~al.(2017)Xun, Li, Zhao, Gao, and
  Zhang}]{xunCorrelatedTopicModel2017}
Guangxu Xun, Yaliang Li, Wayne~Xin Zhao, Jing Gao, and Aidong Zhang. 2017.
\newblock A correlated topic model using word embeddings.
\newblock In \emph{Proceedings of the 26th International Joint Conference on
  Artificial Intelligence}, {{IJCAI}}'17, pages 4207--4213, {Melbourne,
  Australia}. {AAAI Press}.

\bibitem[{Yang et~al.(2017)Yang, Boyd-Graber, and Resnik}]{yang2017adapting}
Weiwei Yang, Jordan Boyd-Graber, and Philip Resnik. 2017.
\newblock Adapting topic models using lexical associations with tree priors.
\newblock In \emph{Proceedings of the 2017 Conference on Empirical Methods in
  Natural Language Processing}, pages 1901--1906.

\bibitem[{Yang et~al.(2020)Yang, Cui, Chen, Che, Liu, Wang, and
  Hu}]{yangTextBrewerOpenSourceKnowledge2020}
Ziqing Yang, Yiming Cui, Zhipeng Chen, Wanxiang Che, Ting Liu, Shijin Wang, and
  Guoping Hu. 2020.
\newblock \href {http://arxiv.org/abs/2002.12620} {{{TextBrewer}}: {{An
  Open}}-{{Source Knowledge Distillation Toolkit}} for {{Natural Language
  Processing}}}.
\newblock \emph{ACL Demo Session}.

\end{thebibliography}
\bibliographystyle{acl_natbib}

\clearpage
\appendix
\section*{Appendix} 

\section{Dev Set Results}\label{appendix:test}

We optimized our models on the dev set, froze the optimal models, and showed the results on the test set in \cref{tab:results_main,tab:results_wlda_kd}. We show the corresponding dev set results for those models in \cref{tab:results_test_main,tab:results_test_wlda_kd}.

\begin{table*}[bhtp]
    \centering
    \resizebox{\textwidth}{!}{
        \begin{tabular}{@{}llllclll@{}}
            \toprule
            \multicolumn{1}{c}{\textbf{}} & \multicolumn{3}{c}{$K = 50$}      &                                   & \multicolumn{3}{c}{$K = 200$}                                                                                                                    \\ %
            \multicolumn{1}{c}{}          & \multicolumn{1}{c}{\textbf{20NG}} & \multicolumn{1}{c}{\textbf{Wiki}} & \multicolumn{1}{c}{\textbf{IMDb}} &  & \multicolumn{1}{c}{\textbf{20NG}} & \multicolumn{1}{c}{\textbf{Wiki}} & \multicolumn{1}{c}{\textbf{IMDb}} \\ \midrule
            \textbf{DVAE}                 & 0.341                             & 0.512                             & 0.137                             &  & 0.312                             & 0.470                             & 0.155                             \\
            \textbf{W-LDA}                & 0.294                             & 0.500                             & 0.136                             &  & 0.203                             & 0.310                             & 0.095                             \\
            \textbf{\textsc{Scholar}}     & 0.343 (0.003)                     & 0.504 (0.007)                     & 0.167 (0.002)                     &  & 0.279 (0.002)                     & 0.478 (0.005)                     & 0.139 (0.002)                     \\ \midrule
            \textbf{\textsc{Sch. + BAT}}  & 0.377 (0.006)                     & 0.526 (0.009)                     & 0.180 (0.002)                     &  & 0.343 (0.002)                     & 0.518 (0.001)                     & 0.174 (0.003)                     \\ \bottomrule
        \end{tabular}
    }
    \caption{The development-set NPMI for our baselines (\cref{sec:baselines}) compared with \ourmodel (explained in \cref{sec:methodology}) using \textsc{Scholar} as our base neural architecture. We achieve better NPMI than all baselines across three datasets and $K=50, K=200$ topics. %
        We use 5 random restarts report the standard deviation.
    }
    \label{tab:results_test_main}
\end{table*}

\begin{table}[hbtp]
    \centering
    \resizebox{\columnwidth}{!}{
        \begin{tabular}{@{}rrrr@{}}
            \toprule
                                & \multicolumn{1}{c}{\textbf{20NG}} & \multicolumn{1}{c}{\textbf{Wiki}} & \multicolumn{1}{c}{\textbf{IMDb}} \\ \midrule
            \textbf{W-LDA}      & 0.294 (0.014)                     & 0.500 (0.013)                     & 0.136 (0.009)                     \\
            \textbf{+\ourmodel} & \textbf{0.316} (0.010)            & \textbf{0.511} (0.016)            & \textbf{0.162} (0.003)            \\ \bottomrule
        \end{tabular}}
    \caption{The mean development-set NPMI (std. dev.) across 5 runs for W-LDA and W-LDA+\ourmodel  for $K=50$, showing improvement on all datasets. This demonstrates that our innovation is \textit{modular} and can be used with base neural topic models that vary in architecture.
    }
    \label{tab:results_test_wlda_kd}
\end{table}
\section{Extrinsic Classification Results}\label{appendix:classification}

The primary goal of our method is to improve the coherence of generated topics. It is natural, however, to ask about the impact of our method on downstream applications. We include here a preliminary exploration suggesting that the addition of BAT does not hurt performance in document classification.

In our setup, we seek to predict document labels $y_d$ from the MAP estimate of a document's topic distribution, $\bm{\theta}_d$. Specifically, we classify the newsgroup to which a document was posted for the 20 newsgroups data (e.g., \texttt{talk.politics.misc}) and a binary sentiment label for the IMDb review data. We train a random forest classifier using default parameters from \texttt{scikit-learn} \cite{scikit-learn} and report the accuracies in \cref{tab:results_classification} (averaged across 5 runs).

Much like other work that is aimed at topic coherence rather than their downstream use in supervised models \cite{nan-etal-2019-topic}, we find that our method has little impact on predictive performance. While it is possible that improvements may be obtained by specifically tuning models for classification, or by integrating BAT into model variations that combine lexical and topic representations \cite[e.g.][]{nguyen2013lexical}, we leave this to future work.

\begin{table}[hbtp]
    \centering
    \begin{tabular}{r r l l }
        \toprule
                      & $K$ & \textbf{\textsc{Scholar}} & +\textbf{\ourmodel} \\
        \midrule
        \texttt{20ng} & 50  & 0.676 (0.003)             & 0.669 (0.005)       \\
                      & 200 & 0.683 (0.002)             & 0.679 (0.004)       \\

        \texttt{IMDb} & 50  & 0.829 (0.003)             & 0.823 (0.011)       \\
                      & 200 & 0.805 (0.003)             & 0.814 (0.004)       \\
        \midrule
    \end{tabular}
    \caption{Random forest classification accuracy on \texttt{20ng} and \texttt{IMDb} datasets, using topic estimates from \textsc{Scholar} and \textsc{Scholar} + \ourmodel.
    }
    \label{tab:results_classification}
\end{table}
\section{Using BAT with DVAE}\label{appendix:dvae}

We further illustrate our method's modularity by applying \textsc{BAT} to our own reimplementation of DVAE \cite{burkhardtDecouplingSparsitySmoothness2019}.\footnote{We appreciate a reviewer's suggestion that we add a +BAT comparison for DVAE.} In contrast to the author's primary implementation, which estimates the model with rejection sampling variational inference (used in \cref{sec:results}), we reimplemented DVAE, approximating the Dirichlet gradient via pathwise derivatives \cite{jankowiakPathwiseDerivativesReparameterization2018}, similar to \citet{burkhardtDecouplingSparsitySmoothness2019}'s alternative model variant using implicit gradients.

Our reimplementation shows baseline behavior substantially similar to the author's implementation. In the course of our experimentation, we noted a degeneracy in this model, in which high NPMI is achieved but at the cost of redundant topics. This failure mode is well-established, but as discussed in \cref{appendix:tu}, we find the measures proposed to diagnose topic diversity \citep[including those proposed by ][]{burkhardtDecouplingSparsitySmoothness2019,nan-etal-2019-topic} to be problematic. Rather than use these metrics, therefore, we took a coarse but simple approach and filtered out any models that yielded more than one pair of identical topics, averaged across five runs (defined as having two topics with the same set of top-10 words). This filtering eliminated many hyperparameter settings, leading us to believe that DVAE is not robust to this problem.

Ultimately, we find that applying \textsc{BAT} to DVAE does not hurt, and also does not help appreciably
(\cref{tab:results_dvae_kd}). In addition, when applying the above filtering criterion to our main \textsc{Scholar} and \textsc{Scholar + BAT} models, we still obtain the positive results reported in \cref{tab:results_test_main}. \footnote{For $K=50$. The single-pair threshold proves too restrictive for the $K=200$ case, where no hyperparameter settings pass the threshold. Increasing the tolerance to a maximum of 5 redundant pairs with $K=200$ leads to a somewhat lower average NPMI overall, but the same directional improvement, i.e. \textsc{Scholar+BAT} yields a significantly higher NPMI than \textsc{Scholar}.}

\begin{table}[hbtp]
    \centering
    \resizebox{\columnwidth}{!}{
        \begin{tabular}{@{}rrrr@{}}
            \toprule
                          & \multicolumn{1}{c}{\textbf{20NG}} & \multicolumn{1}{c}{\textbf{Wiki}} & \multicolumn{1}{c}{\textbf{IMDb}} \\ \midrule
            \textbf{DVAE} & 0.376 (0.004)                     & \textbf{0.517} (0.006)            & \textbf{0.169} (0.007)            \\
            \textbf{+BAT} & \textbf{0.401} (0.005)            & \textbf{0.515} (0.007)            & \textbf{0.169} (0.006)            \\\bottomrule
        \end{tabular}}
    \caption{
        Mean development set NPMI (s.d.) across 5 runs for DVAE \cite{burkhardtDecouplingSparsitySmoothness2019} and DVAE+\ourmodel for $K=50$.
    }
    \label{tab:results_dvae_kd}
\end{table}
\section{Methodological Notes}
\subsection{Using BERT in the encoder}
\label{appendix:encoderbert}

In \textsc{Scholar}, the encoder takes the following form:
\begin{align}
    \bm{\pi}_d    & = g\left(\left[\bm{W} \bm{w}^\textsc{BoW}_d \right]\right)                 \\
    \bm{\theta}_d & \sim \mathcal{LN}\left(\mu_\nu(\bm{\pi}_d),\,\sigma_\nu(\bm{\pi}_d)\right)
\end{align}
Where the weight matrix $\bm{W}$, along with the parameters of nueral networks $\mu(\cdot)$ and $\sigma(\cdot)$, are our variational parameters.

\newcite{card-etal-2018-neural} propose that pre-trained word2vec \cite{mikolovEfficientEstimationWord2013} embeddings can replace $\bm{W}$, meaning that the document representation made available to the encoder is an $l$-dimensional sum of word embeddings. \newcite{card-etal-2018-neural} argue that fixed embeddings act as an inductive prior which improves topic coherence. Likewise, we might want to encode the document representation from a BERT-like model and, in fact, this has been attempted with some success \cite{bianchiPretrainingHotTopic2020}. The hypothesis is that a structure-dependent  representation of the document can better parameterize its corresponding topic distribution.

We experimented with this method as well, using both the hidden BERT representation and the predicted probabilities, although we also include a fixed randomized baseline to maintain parameter parity. Results for IMDb are reported in \cref{tab:encoderbert}, and we find at best a mild improvement over the baselines.\footnote{We also fail to reproduce the findings of \newcite{card-etal-2018-neural}, showing no meaningful improvement in topic coherence with fixed word2vec embeddings. It appears that this is a consequence of their tuning for perplexity rather than NPMI.} We suspect the reason for this tepid result is both that (a) in training, the effect of estimated local document-topic proportions on the global topic-word distributions is diffuse and indirect; and (b) the compression of the representation into $k$ dimensions causes too much of the high-level linguistic information to be lost. Nonetheless, owing to the slight benefit, we do pass the logits to the encoder in our \textsc{Scholar}-based model. We avoid this change for the model based on W-LDA to underscore the modularity of our method.

\begin{table}[]
    \footnotesize
    \centering
    \begin{tabular}{l r}
        \toprule
        \textbf{Setting}                               & \textbf{NPMI} \\
        \midrule
        Randomly updated embeds.                       & 0.170 (0.007) \\
        Fixed word2vec embeds.                         & 0.172 (0.004) \\
        Random 784-dim doc. rep. + w2v                 & 0.175 (0.007) \\
        Mean-pooled 784-dim BERT output          + w2v & 0.172 (0.002) \\
        Random 5000-dim doc. rep. + w2v                & 0.178 (0.007) \\
        5000-dim predicted probs. from BAT + w2v       & 0.180 (0.008) \\
        \bottomrule
    \end{tabular}
    \caption{
        Effect on topic coherence of passing various document representations to the \textsc{Scholar} encoder (using the IMDb data). Each setting describes the document representation provided to the encoder, which is transformed by one feed-forward layer of 300-dimensions followed by a second down to $K$ dimensions. ``+ w2v'' indicates that we first concatenated with the sum of the 300-dimensional word2vec embeddings for the document.
        Note that these early findings are based on a different IMDb development set, a 20\% split from the training data. They are thus not directly comparable to the results reported elsewhere in the text, which used a separate held-out development set.
    }
    \label{tab:encoderbert}
\end{table}

\subsection{Topic Diversity}\label{appendix:tu}

\newcite{burkhardtDecouplingSparsitySmoothness2019} have found a degeneracy in some topic models, wherein a single topic will be repeated more than once with slightly varying terms (e.g., several \underline{Dadaism} topics). 
\newcite{burkhardtDecouplingSparsitySmoothness2019} and others \cite{nan-etal-2019-topic,diengTopicModelingEmbedding2019} have independently proposed related metrics to quantify the problem, but the literature has not converged on a solution. In contrast to NPMI, we are not aware of any work that assesses the validity of such metrics with respect to human judgements.

Moreover, all these proposals suffer from a common problem: because they are global measures of word overlap, they fail to account for \emph{how} words are repeated across topics. For instance, Topic Uniqueness \cite{nan-etal-2019-topic} is identical regardless of whether all of a topic's top words are all repeated in a single second topic, or individual top words from that topic are repeated in several other topics. In addition, the measures inappropriately penalize partially-related topics.

They also penalize polysemy---and, more generally, the contextual flexibility of word meanings.
One of the key \emph{advantages} of latent topics, compared to surface lexical summaries, is that the same word can contribute differently to an understanding of what different topics are about. As a real example from our experience, in modeling a set of documents related to paid family and medical leave, words like \emph{parent}, \emph{mother}, and \emph{father} are prominent in one topic related to parental leave when a child is born (accompanying other terms like \emph{newborn} and \emph{maternity\_leave}) and also in another topic related to taking leave to care for family members, including elderly parents  (accompanying other terms like \emph{elderly} and \emph{aging}). The fact that topic models permit a word like \emph{parent} to be prominent in both of these clearly distinct topics, emphasizing two different aspects of the word relative to the collection as a whole (being a parent taking care of children, being a child taking care of parents), is a feature, not a bug. We consider the question of topic diversity an important direction for future work.

\section{Experimental Procedures}
In this section, we first provide details of our hyperparameters and tuning procedures, then turn to our computing infrastructure and the rough runtime of the \textsc{Scholar} model.

\subsection{Hyperparameter Tuning and Optimal Values}\label{appendix:hyperparam}

We used well-tuned baselines to establish thresholds for performance on NPMI \citep[following the reported hyperparameters in][]{card-etal-2018-neural, burkhardtDecouplingSparsitySmoothness2019, nan-etal-2019-topic}. While developing our model, we performed a coarse-grained initial hyperparameter sweep to identify ranges that were not beating the threshold, and decided to exclude those ranges when performing a full grid search.
We report the hyperparameter ranges used in this search, along with their optimal values (as determined by development set NPMI), in \cref{tab:hyperparam_scholar_20ng,tab:hyperparam_scholar_wiki,tab:hyperparam_scholar_imdb,tab:hyperparam_wlda,tab:hyperparam_dvae}. These produced the final set of results (\cref{tab:results_main,tab:results_wlda_kd,tab:results_test_main,tab:results_test_wlda_kd}).

For the \textsc{DistilBERT} training, we use the default hyperparameter settings for the \texttt{bert-base-uncased} model \cite{Wolf2019HuggingFacesTS}. Our code is a modified version of the \textsc{MM-IMDb} multimodal sequence classification code from the same codebase as \textsc{DistilBERT} ({\url{https://github.com/huggingface/transformers/tree/master/examples/contrib/mm-imdb}), and we use all default hyperparameter settings specified there. We train for 7500 steps for \texttt{20ng}, and 17000 steps for \texttt{Wiki} and \texttt{IMDb} (this corresponds to convergence on development-set perplexity).

\begin{table*}[]
    \centering
    \scalebox{0.75}{
        \begin{tabular}{@{}lrrrrr@{}}
            \toprule
            \multicolumn{2}{c}{Dataset: 20NG} & \multicolumn{2}{c}{k = 50}                & \multicolumn{2}{c}{k = 200}                                                                                                                                                                                                       \\ \midrule
            \textbf{}                         & \multicolumn{1}{c}{\textbf{Values Tried}} & \multicolumn{1}{c}{\textbf{\begin{tabular}[c]{@{}c@{}}SCHOLAR\\ (optimal values)\end{tabular}}} & \multicolumn{1}{c}{\textbf{\begin{tabular}[c]{@{}c@{}}SCHOLAR+BAT\\ (optimal values)\end{tabular}}} & \multicolumn{1}{c}{\textbf{\begin{tabular}[c]{@{}c@{}}SCHOLAR\\ (optimal values)\end{tabular}}} & \multicolumn{1}{c}{\textbf{\begin{tabular}[c]{@{}c@{}}SCHOLAR+BAT\\ (optimal values)\end{tabular}}} \\ \midrule
            \textbf{lr}                       & 0.002*                                    & 0.002                                                  & 0.002                                                  & 0.002                                                  & 0.002                                                  \\
            \textbf{$\bm{\alpha}$}            & 1.0*                                      & 1.0                                                    & 1.0                                                    & 1.0                                                    & 1.0                                                    \\
            \textbf{$\bm{\lambda}$}           & $\{0.25, 0.5, 0.75, 0.95, 0.99, 0.999\}$  & -                                                      & 0.75                                                   & -                                                      & 0.99                                                   \\
            \textbf{$\bm{T}$}                 & $\{1.0, 2.0, 3.0, 5.0\}$                  & -                                                      & 2.0                                                    & -                                                      & 5.0                                                    \\ \bottomrule
        \end{tabular}}
    \caption{Hyperparameter ranges and optimal values (as determined by development set NPMI) for \textsc{Scholar} and \textsc{Scholar}+\ourmodel, on the \textbf{20NG} dataset. \textbf{lr} is the learning rate, \textbf{$\bm{\alpha}$} is the hyperparameter for the logistic normal prior, \textbf{$\bm{\lambda}$} is the weight on the teacher model logits from \cref{eq:kd-loss}, and \textbf{$\bm{T}$} is the softmax temperature from \cref{eq:kd-loss}. Other hyperparamter values (which can be accessed in our code base) which were kept at their default values are not reported here. Values marked with the * are also kept at their default values per the base SCHOLAR model (\url{https://github.com/dallascard/scholar}). All different sweeps in the grid search were run for \textbf{500 epochs} with a \textbf{batch size = 200}. }
    \label{tab:hyperparam_scholar_20ng}
\end{table*}

\begin{table*}[]
    \centering
    \scalebox{0.75}{
        \begin{tabular}{@{}lrrrrr@{}}
            \toprule
            \multicolumn{2}{c}{Dataset: Wiki} & \multicolumn{2}{c}{k = 50}                & \multicolumn{2}{c}{k = 200}                                                                                                                                                                                                       \\ \midrule
            \textbf{}                         & \multicolumn{1}{c}{\textbf{Values Tried}} & \multicolumn{1}{c}{\textbf{\begin{tabular}[c]{@{}c@{}}SCHOLAR\\ (optimal values)\end{tabular}}} & \multicolumn{1}{c}{\textbf{\begin{tabular}[c]{@{}c@{}}SCHOLAR+BAT\\ (optimal values)\end{tabular}}} & \multicolumn{1}{c}{\textbf{\begin{tabular}[c]{@{}c@{}}SCHOLAR\\ (optimal values)\end{tabular}}} & \multicolumn{1}{c}{\textbf{\begin{tabular}[c]{@{}c@{}}SCHOLAR+BAT\\ (optimal values)\end{tabular}}} \\ \midrule
            \textbf{lr}                       & $\{0.001, 0.002, 0.005\}$                 & 0.001                                                  & 0.001                                                  & 0.002                                                  & 0.005                                                  \\
            \textbf{$\bm{\alpha}$}            & $\{0.0005, 0.00075, 0.001, 0.005, 0.01\}$ & 0.01                                                   & 0.00075                                                & 0.0005                                                 & 0.001                                                  \\
            \textbf{anneal}                   & $\{0.25, 0.5, 0.75\}$                     & 0.25                                                   & 0.5                                                    & 0.25                                                   & 0.5                                                    \\
            \textbf{$\bm{\lambda}$}           & $\{0.4, 0.5, 0.6, 0.7, 0.75, 0.8\}$       & -                                                      & 0.75                                                   & -                                                      & 0.75                                                   \\
            \textbf{$\bm{T}$}                 & $\{1.0, 2.0\}$                            & -                                                      & 1.0                                                    & -                                                      & 1.0                                                    \\
            \textbf{clipping}                 & $\{1.0, 1.5, 2.0\}$                       & -                                                      & 2.0                                                    & -                                                      & 1.5                                                    \\ \bottomrule
        \end{tabular}}
    \caption{Hyperparameter ranges and optimal values (as determined by development set NPMI) for \textsc{Scholar} and \textsc{Scholar}+\ourmodel, on the \textbf{Wiki} dataset. \textbf{lr} is the learning rate, \textbf{$\bm{\alpha}$} is the hyperparameter for the logistic normal prior, \textbf{anneal} controls the annealing (as explained in Appendix B in \newcite{card-etal-2018-neural}), \textbf{$\bm{\lambda}$} is the weight on the teacher model logits from \cref{eq:kd-loss}, \textbf{$\bm{T}$} is the softmax temperature from \cref{eq:kd-loss}, and \textbf{clipping} controls how much of the logit distribution to clip (\cref{sec:methodology}). Other hyperparamter values (which can be accessed in our code base) which were kept at their default values are not reported here. All different sweeps in the grid search were run for \textbf{500 epochs} with a \textbf{batch size = 500}. }
    \label{tab:hyperparam_scholar_wiki}
\end{table*}

\begin{table*}[]
    \centering
    \scalebox{0.75}{
        \begin{tabular}{@{}lrrrrr@{}}
            \toprule
            \multicolumn{2}{c}{Dataset: IMDb} & \multicolumn{2}{c}{k = 50}                & \multicolumn{2}{c}{k = 200}                                                                                                                                                                                                          \\ \midrule
            \textbf{}                         & \multicolumn{1}{c}{\textbf{Values Tried}} & \multicolumn{1}{c}{\textbf{\begin{tabular}[c]{@{}c@{}}SCHOLAR\\ (optimal values)\end{tabular}}} & \multicolumn{1}{c}{\textbf{\begin{tabular}[c]{@{}c@{}}SCHOLAR+BAT\\ (optimal values)\end{tabular}}} & \multicolumn{1}{c}{\textbf{\begin{tabular}[c]{@{}c@{}}SCHOLAR\\ (optimal values)\end{tabular}}} & \multicolumn{1}{c}{\textbf{\begin{tabular}[c]{@{}c@{}}SCHOLAR+BAT\\ (optimal values)\end{tabular}}} \\ \midrule
            \textbf{lr}                       & 0.002*                                    & 0.002                                                  & 0.002                                                   & 0.002                                                   & 0.002                                                   \\
            \textbf{$\bm{\alpha}$}            & $\{0.01, 0.1, 0.5, 1.0\}$                 & 0.5                                                    & 0.5                                                     & 0.1                                                     & 0.1                                                     \\
            \textbf{anneal}                   & $\{0.25, 0.5, 0.75\}$                     & 0.25                                                   & 0.25                                                    & 0.25                                                    & 0.5                                                     \\
            \textbf{$\bm{\lambda}$}           & $\{0.25, 0.5, 0.75, 0.99\}$               & -                                                      & 0.5                                                     & -                                                       & 0.99                                                    \\
            \textbf{$\bm{T}$}                 & $\{1.0, 2.0\}$                            & -                                                      & 1.0                                                     & -                                                       & 1.0                                                     \\
            \textbf{clipping}                 & $\{0.0, 1.0, 10.0\}$                      & -                                                      & 10.0                                                    & -                                                       & 0.0                                                     \\ \bottomrule
        \end{tabular}}
    \caption{Hyperparameter ranges and optimal values (as determined by development set NPMI) for \textsc{Scholar} and \textsc{Scholar}+\ourmodel, on the \textbf{IMDb} dataset. \textbf{lr} is the learning rate, \textbf{$\bm{\alpha}$} is the hyperparameter for the logistic normal prior, \textbf{anneal} controls the annealing (as explained in Appendix B in \newcite{card-etal-2018-neural}), \textbf{$\bm{\lambda}$} is the weight on the teacher model logits from \cref{eq:kd-loss}, \textbf{$\bm{T}$} is the softmax temperature from \cref{eq:kd-loss}, and \textbf{clipping} controls how much of the logit distribution to clip (\cref{sec:methodology}). Other hyperparamter values (which can be accessed in our code base) which were kept at their default values are not reported here. Values marked with the * are also kept at their default values per the base SCHOLAR model (\url{https://github.com/dallascard/scholar}). All different sweeps in the grid search were run for \textbf{500 epochs} with a \textbf{batch size = 200}. }
    \label{tab:hyperparam_scholar_imdb}
\end{table*}

\begin{table*}[]
    \centering
    \begin{tabular}{@{}lrrr@{}}
        \toprule
        \multicolumn{2}{c}{(Dataset: 20NG)} & \multicolumn{2}{c}{}                                                                                                                                          \\ \midrule
        \textbf{}                           & \multicolumn{1}{c}{\textbf{Values Tried}} & \multicolumn{1}{c}{\textbf{\begin{tabular}[c]{@{}c@{}}W-LDA\\ (optimal values)\end{tabular}}} & \multicolumn{1}{c}{\textbf{\begin{tabular}[c]{@{}c@{}}W-LDA+\ourmodel\\ (optimal values)\end{tabular}}} \\
        \textbf{lr}                         & $\{0.002\}$                               & 0.002                                                   & 0.002                                                   \\
        \textbf{$\bm{\alpha}$}              & $\{0.1, 1.0\}$                            & 0.1                                                     & 0.1                                                     \\
        \textbf{$\bm{\lambda}$}             & $\{0.75, 0.99\}$                          & -                                                       & 0.75                                                    \\
        \textbf{$\bm{T}$}                   & $\{1.0, 2.0\}$                            & -                                                       & 1.0                                                     \\ \midrule
        \multicolumn{2}{c}{(Dataset: Wiki)} &                                           &                                                                                                                   \\ \midrule
        \textbf{lr}                         & $\{0.001\}$                               & 0.001                                                   & 0.001                                                   \\
        \textbf{$\bm{\alpha}$}              & $\{0.01, 0.1\}$                           & 0.1                                                     & 0.1                                                     \\
        \textbf{$\bm{\lambda}$}             & $\{0.25, 0.75\}$                          & -                                                       & 0.25                                                    \\
        \textbf{$\bm{T}$}                   & $\{1.0, 2.0, 5.0\}$                       & -                                                       & 2.0                                                     \\
        \textbf{clipping}                   & $\{1.0, 2.0\}$                            & -                                                       & 1.0                                                     \\ \midrule
        \multicolumn{2}{c}{(Dataset: IMDb)} &                                           &                                                                                                                   \\ \midrule
        \textbf{lr}                         & $\{0.002\}$                               & 0.002                                                   & 0.002                                                   \\
        \textbf{$\bm{\alpha}$}              & $\{0.1\}$                                 & 0.1                                                     & 0.1                                                     \\
        \textbf{$\bm{\lambda}$}             & $\{0.75\}$                                & -                                                       & 0.75                                                    \\
        \textbf{$\bm{T}$}                   & $\{1.0\}$                                 & -                                                       & 1.0                                                     \\ \bottomrule
    \end{tabular}
    \caption{Hyperparameter ranges and optimal values (as determined by development set NPMI) for \textsc{W-LDA} and \textsc{W-LDA}+\ourmodel, on all three datasets. \textbf{lr} is the learning rate, \textbf{$\bm{\alpha}$} is the hyperparameter for the dirichlet prior, \textbf{$\bm{\lambda}$} is the weight on the teacher model logits from \cref{eq:kd-loss}, \textbf{$\bm{T}$} is the softmax temperature from \cref{eq:kd-loss}, and \textbf{clipping} controls how much of the logit distribution to clip (\cref{sec:methodology}). Other hyperparameter values (which can be accessed in our codebase) which were kept at their default values in the original baseline code are not reported here (also see \newcite{nan-etal-2019-topic} and \url{https://github.com/awslabs/w-lda/}). Values marked with the * are also kept at their default values. All different sweeps in the grid search were run for \textbf{500 epochs} and noise parameter = 0.5 (see \newcite{nan-etal-2019-topic}). For 20NG and IMDb, we used \textbf{batch size = 200}, and for Wiki, we used \textbf{batch size = 360}.}
    \label{tab:hyperparam_wlda}
\end{table*}

\begin{table*}[]
    \centering
    \begin{tabular}{@{}ccccccc@{}}
        \toprule
        \textbf{}                                            & \multicolumn{3}{c}{k = 50} & \multicolumn{3}{c}{k = 200}                                                                                                      \\ \midrule
                                                             & \textbf{20NG}              & \textbf{Wiki}               & \textbf{IMDb}           & \textbf{20NG}           & \textbf{Wiki}        & \textbf{IMDb}           \\ \midrule
        \multicolumn{1}{l}{\textbf{Optimal Dirichlet Prior}} & \multicolumn{1}{r}{0.6}    & \multicolumn{1}{r}{}        & \multicolumn{1}{r}{0.2} & \multicolumn{1}{r}{0.6} & \multicolumn{1}{r}{} & \multicolumn{1}{r}{0.2} \\ \bottomrule
    \end{tabular}
    \caption{For DVAE, we tried four values for the Dirichlet Prior (as per the values tried by the authors in \newcite{burkhardtDecouplingSparsitySmoothness2019}) - \textbf{$\{0.01, 0.1, 0.2, 0.6\}$} and report the optimal values corresponding to the dev set results (\cref{tab:results_main}) and test set results (\cref{tab:results_test_main}) in this table. Within the model variations available in the codebase for DVAE (\url{https://github.com/sophieburkhardt/dirichlet-vae-topic-models}) we use the Dirichlet VAE based on RSVI which is shown to give the highest NPMI scores in \newcite{burkhardtDecouplingSparsitySmoothness2019}. }
    \label{tab:hyperparam_dvae}
\end{table*}

\subsection{Computing Infrastructure and Runtime}

For the full hyperparameter sweep, we used an Amazon Web Services ParallelCluster \url{https://github.com/aws/aws-parallelcluster} with 40 nodes of \texttt{g4dn.xlarge} instances (consisting of Nvidia T4 GPUs with 16 GB RAM), which ran for about 5 days. For initial experimentation, we used a SLURM cluster with a mix of consumer-grade Nvidia GPUs (e.g., 1080, 2080).

In terms of runtime, \textsc{Scholar}) and our own \textsc{Scholar}+\ourmodel are equal and this is true for any of our baseline model augmented with BAT. It is important to note that the overhead in terms of the overall runtime comes only from training the \textsc{DistilBERT} encoder on the full dataset first and inference time for obtaining the logits after training. Thus, users should keep in mind the initial step of training and inferring teacher model logits and saving them; once that is done for the dataset, our model does not add to the runtime. We show the comparison between the full runtimes, including the initial step, in \cref{fig:runtime_comparison}. %

\begin{figure*}[t!]
    \centering
    \includegraphics[scale=0.55]{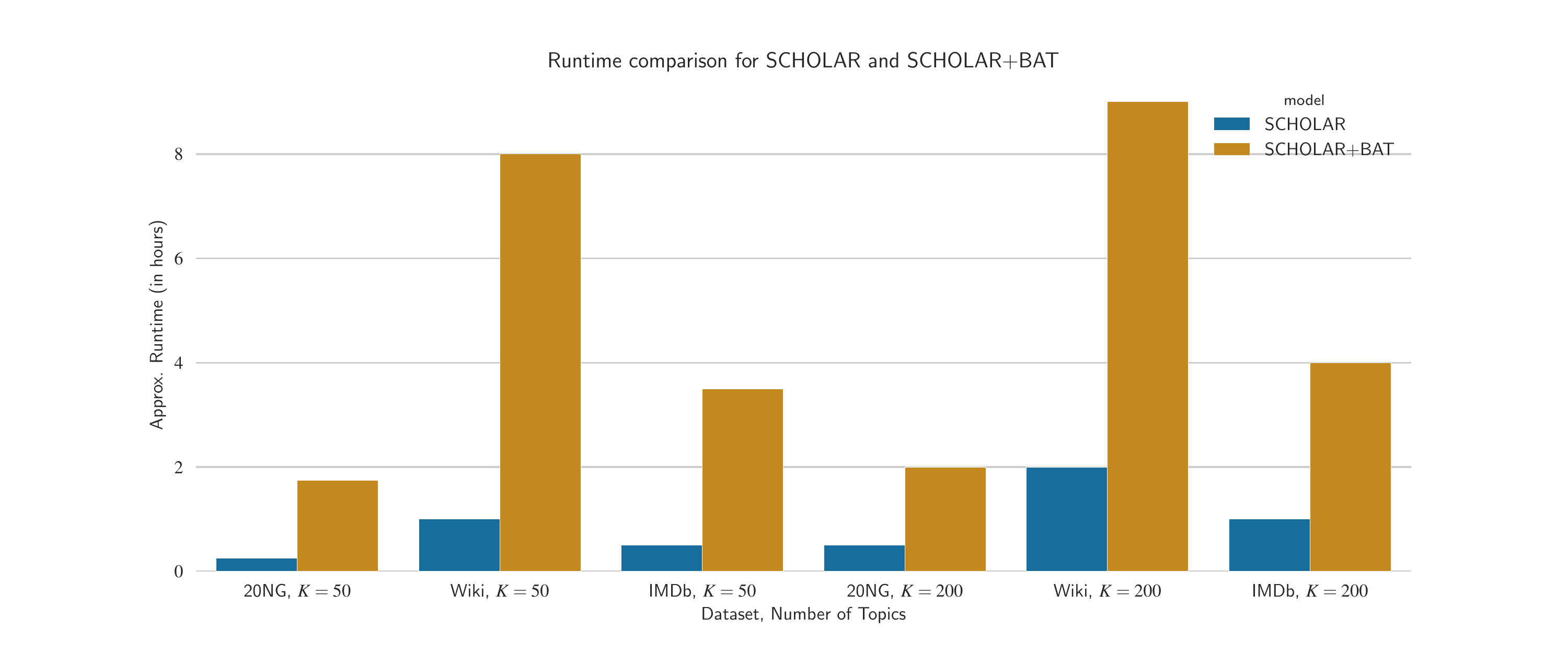}
    \caption{Runtime comparison for \textsc{Scholar}) and our own \textsc{Scholar}+\ourmodel - Note that the overhead due to \ourmodel is only due to the training and inference time required to obtain the \textsc{DistilBERT} encoder logits on the full dataset first, and once the teacher logits are available, the run time of both models is the same. We depict the full approximate time (in hours) including this initial overhead in case of \ourmodel. }
    \label{fig:runtime_comparison}
\end{figure*}

\section{Changes to W-LDA}\label{appendix:code}
In \cref{fig:wldacode}, we show the changes to the W-LDA model necessary to accommodate our method. Ignoring the code to load \& clip the logits, also constituting a minor change, we introduce about a dozen lines.

\begin{figure*}[]
\begin{minted}[fontsize=\footnotesize]{python}
    ### In `compute_op.py`

    ## Retrieve BERT logits
    docs = self.data.get_documents(key='train')
    if self.args['use_kd']:
        split_on = docs.shape[1] // 2
        docs, bert_logits = docs[:,:split_on], docs[:,split_on:]
        t = self.args['kd_softmax_temp']
        kd_docs = nd.softmax(bert_logits / t) * nd.sum(docs, axis=1, keepdims=True)

    # [... unchanged lines ...]

    ## Compute loss
    with autograd.record():
        # [... unchanged lines ...]
        if self.args['use_kd']:
            kd_logits = nd.log_softmax(x_reconstruction_u / t)
            logits = nd.log_softmax(x_reconstruction_u)

            kd_loss_reconstruction = nd.mean(nd.sum(- kd_docs * kd_logits, axis=1))
            loss_reconstruction = nd.mean(nd.sum(- docs * logits, axis=1))

            loss_total = self.args['recon_alpha'] * (
                self.args['kd_loss_alpha'] * t * t * (kd_loss_reconstruction) +
                (1 - self.args['kd_loss_alpha']) * loss_reconstruction
            )
        else: 
            # [... unchanged lines ...]
\end{minted}
\caption{
    Modified portions of W-LDA model to accommodate BAT. We omit definitions of additional command-line arguments and data loading, but they are similarly brief.
}
\label{fig:wldacode}
\end{figure*}
\section{Impact of BAT on Individual Topics: Aligned Topic Pair Examples}\label{appendix:analysis}

For each corpus (20NG, Wiki, and IMDb), a single comparison of base and BAT-augmented (SCHOLAR vs. SCHOLAR+BAT) 50-topic models was selected randomly, from the five runs used in computing average performance in \cref{fig:bar_h2h_wins}.

For each of those pairs of models, we then randomly selected 15~aligned topic pairs from that set of 50 to include in the tables below.  Specifically, a full set of 50~topic pairs was partitioned according to JS divergence into the 10 most similar pairs, the next 10 most similar, and so forth, for a total of five ``brackets'' of topic alignment quality. Three topic pairs were then selected at random from each bracket, hence 15 pairs in all, in order to yield a fair picture of what pairs look like at various qualities of topic alignment.

In the tables below (\cref{tab:all_topic_pairs_20ng,tab:all_topic_pairs_wiki,tab:all_topic_pairs_imdb}), we present pairs sorted from best to worst alignment quality. Recall that for NPMI, higher is better, and for JS divergence, lower score indicates a higher quality match (or alignment) for the topic pair.

\begin{table*}[]
    \centering
    \footnotesize
    \scalebox{0.77}{
        \begin{tabular}{@{}rlr@{}}
            \toprule
            \multicolumn{1}{c}{\textbf{Pair \#}} & \multicolumn{1}{c}{\textbf{\begin{tabular}[c]{@{}c@{}}SCHOLAR vs SCHOLAR+BAT\\ (NPMI, Top 10 Topic Words)\end{tabular}}} & \multicolumn{1}{c}{\textbf{JS Divergence}} \\ \midrule
            1                                    & \begin{tabular}[c]{@{}l@{}}SCHOLAR: (0.399, 'sin eternal lord heaven pray christ prayer jesus god hell')\\ SCHOLAR+BAT: (0.394, 'eternal god hell sin heaven christ jesus christianity faith life')\end{tabular}                              & 0.0287                                     \\ \midrule
            4                                    & \begin{tabular}[c]{@{}l@{}}SCHOLAR: (0.3512, 'score goal puck penalty season shot tie pitch game defensive')\\ SCHOLAR+BAT: (0.3838, 'score goal season game puck shot leafs penalty play playoff')\end{tabular}                              & 0.0345                                     \\ \midrule
            8                                    & \begin{tabular}[c]{@{}l@{}}SCHOLAR: (0.4307, 'doctrine church catholic scripture spirit biblical revelation bible resurrection christ')\\ SCHOLAR+BAT: (0.4454, 'biblical church bible scripture doctrine catholic interpretation passage teaching jesus')\end{tabular}                              & 0.0417                                     \\ \midrule
            11                                   & \begin{tabular}[c]{@{}l@{}}SCHOLAR: (0.7109, 'turks armenian genocide jews mountain armenians turkish proceed nazi armenia')\\ SCHOLAR+BAT: (0.7297, 'turks genocide turkish armenian armenia armenians massacre turkey proceed muslim')\end{tabular}                              & 0.0425                                     \\ \midrule
            15                                   & \begin{tabular}[c]{@{}l@{}}SCHOLAR: (0.2626, 'cryptography security network privacy mailing internet mail encrypt anonymous user')\\ SCHOLAR+BAT: (0.289, 'anonymous mail network privacy internet security cryptography encrypt electronic ftp')\end{tabular}                              & 0.0479                                     \\ \midrule
            17                                   & \begin{tabular}[c]{@{}l@{}}SCHOLAR: (0.3501, 'rider bike ride helmet motorcycle dog bmw dod honda seat')\\ SCHOLAR+BAT: (0.3843, 'helmet bike rider ride dog motorcycle dod rear honda bmw')\end{tabular}                              & 0.0498                                     \\ \midrule
            22                                   & \begin{tabular}[c]{@{}l@{}}SCHOLAR: (0.307, 'voltage circuit amp heat battery electronics frequency signal audio ac')\\ SCHOLAR+BAT: (0.3236, 'circuit voltage amp wire audio wiring signal outlet input pin')\end{tabular}                              & 0.0641                                     \\ \midrule
            27                                   & \begin{tabular}[c]{@{}l@{}}SCHOLAR: (0.4018, 'passage verse jesus biblical resurrection scripture translation interpretation bible prophet')\\ SCHOLAR+BAT: (0.5262, 'jesus christ lord sin heaven resurrection holy mary father son')\end{tabular}                              & 0.071                                      \\ \midrule
            28                                   & \begin{tabular}[c]{@{}l@{}}SCHOLAR: (0.2469, 'nt printer windows microsoft mac unix postscript pc os print')\\ SCHOLAR+BAT: (0.2786, 'font color image format printer display pixel graphic postscript directory')\end{tabular}                             & 0.0729                                     \\ \midrule
            31                                   & \begin{tabular}[c]{@{}l@{}}SCHOLAR: (0.2109, 'crash backup gateway disk windows install memory boot floppy cache')\\ SCHOLAR+BAT: (0.3141, 'disk floppy dos scsi ram cache controller isa swap windows')\end{tabular}                             & 0.0864                                     \\ \midrule
            35                                   & \begin{tabular}[c]{@{}l@{}}SCHOLAR: (0.2589, 'scientific science disease medicine treatment energy observe observation patient scientist')\\ SCHOLAR+BAT: (0.2705, 'science morality objective scientific moral existence observation universe definition theory')\end{tabular}                             & 0.093                                      \\ \midrule
            40                                   & \begin{tabular}[c]{@{}l@{}}SCHOLAR: (0.1252, 'interested kit sale advance email address australia thanks april mail')\\ SCHOLAR+BAT: (0.206, 'mail email mailing address list thanks interested fax please send')\end{tabular}                             & 0.1173                                     \\ \midrule
            41                                   & \begin{tabular}[c]{@{}l@{}}SCHOLAR: (0.2842, 'insurance tax hospital coverage health pay canadian kid care economy')\\ SCHOLAR+BAT: (0.2354, 'dealer car price insurance buy pay sell money honda ford')\end{tabular}                             & 0.1319                                     \\ \midrule
            45                                   & \begin{tabular}[c]{@{}l@{}}SCHOLAR: (0.3165, 'waco clinton president bush senate batf tax fbi compound vote')\\ SCHOLAR+BAT: (0.5144, 'nsa crypto clipper escrow wiretap secure encryption chip warrant scheme')\end{tabular}                             & 0.1791                                     \\ \midrule
            48                                   & \begin{tabular}[c]{@{}l@{}}SCHOLAR: (0.2329, 'screen mouse monitor printer inch resolution tube apple font print')\\ SCHOLAR+BAT: (0.275, 'heat fuel tube cool detector radar gas nuclear hole cold')\end{tabular}                             & 0.2527                                     \\ \bottomrule
        \end{tabular}}
    \caption{Fifteen aligned topic pairs from the 20NG dataset.}
    
    \label{tab:all_topic_pairs_20ng}
\end{table*}

\begin{table*}[]
    \centering
    \footnotesize
    \scalebox{0.77}{
        \begin{tabular}{@{}rlr@{}}
            \toprule
            \multicolumn{1}{c}{\textbf{Pair \#}} & \multicolumn{1}{c}{\textbf{\begin{tabular}[c]{@{}c@{}}SCHOLAR vs SCHOLAR+BAT\\ (NPMI, Top 10 Topic Words)\end{tabular}}} & \multicolumn{1}{c}{\textbf{JS Divergence}} \\ \midrule
            1                                    & \begin{tabular}[c]{@{}l@{}}SCHOLAR: (0.5804, 'prognosis protein symptom intravenous diagnosis syndrome medication abnormality infection dysfunction')\\ SCHOLAR+BAT: (0.5464, 'abnormality prognosis intravenous receptor syndrome antibiotic inflammation diagnosis mutation dos')\end{tabular}                              & 0.163                                      \\ \midrule
            4                                    & \begin{tabular}[c]{@{}l@{}}SCHOLAR: (0.6036, 'parsec brightest orbiting astronomer planetary brightness luminosity jupiter constellation orbit')\\ SCHOLAR+BAT: (0.586, 'orbiting habitable gliese planetary extrasolar parsec brightness luminosity orbital jupiter')\end{tabular}                              & 0.1787                                     \\ \midrule
            8                                    & \begin{tabular}[c]{@{}l@{}}SCHOLAR: (0.4432, 'lap peloton uci breakaway sprint ferrari bmc tyre podium sauber')\\ SCHOLAR+BAT: (0.4902, 'lap sprint podium finisher quickest uci mclaren ferrari peloton rosberg')\end{tabular}                              & 0.1879                                     \\ \midrule
            11                                   & \begin{tabular}[c]{@{}l@{}}SCHOLAR: (0.5662, 'ny renumbering cr realigned intersects intersecting hamlet concurrency routing truncated')\\ SCHOLAR+BAT: (0.5888, 'ny intersects renumbering intersecting realigned cr concurrency routing intersection hamlet')\end{tabular}                              & 0.1989                                     \\ \midrule
            15                                   & \begin{tabular}[c]{@{}l@{}}SCHOLAR: (0.4866, 'byzantine caliphate ibn caliph byzantium abbasid thrace constantinople vassal umayyad')\\ SCHOLAR+BAT: (0.4686, 'byzantium thrace caliphate nikephoros antioch byzantine envoy umayyad principality constantinople')\end{tabular}                              & 0.2076                                     \\ \midrule
            17                                   & \begin{tabular}[c]{@{}l@{}}SCHOLAR: (0.4944, 'gubernatorial kentucky reelection republican democrat frankfort candidacy legislator congressman caucus')\\ SCHOLAR+BAT: (0.494, 'gubernatorial reelection legislator congressman candidacy caucus whig democrat kentucky veto')\end{tabular}                              & 0.2211                                     \\ \midrule
            22                                   & \begin{tabular}[c]{@{}l@{}}SCHOLAR: (0.4069, 'electrification electrified locomotive train nok railway freight oslo commuter nsb')\\ SCHOLAR+BAT: (0.3567, 'nok electrified electrification commuter oslo tramway freight livery bergen locomotive')\end{tabular}                              & 0.2391                                     \\ \midrule
            27                                   & \begin{tabular}[c]{@{}l@{}}SCHOLAR: (0.4187, 'gatehouse chancel nave stonework anglesey castle demography domesday storey vaulted')\\ SCHOLAR+BAT: (0.4035, 'domesday demography cheshire gatehouse storey borough manor priory mersey avon')\end{tabular}                              & 0.2564                                     \\ \midrule
            28                                   & \begin{tabular}[c]{@{}l@{}}SCHOLAR: (0.4041, 'frigate brig convoy hm torpedoed rigging destroyer sailed sighted starboard')\\ SCHOLAR+BAT: (0.4126, 'brig frigate privateer rigging schooner sloop corvette sighted indiaman brest')\end{tabular}                              & 0.2617                                     \\ \midrule
            31                                   & \begin{tabular}[c]{@{}l@{}}SCHOLAR: (0.2876, 'raaf battalion aircrew beachhead moresby amberley brigade usaaf dso jagdgeschwader')\\ SCHOLAR+BAT: (0.5148, 'platoon counterattack bridgehead divisional battalion mortar perimeter brigade beachhead regimental')\end{tabular}                              & 0.2651                                     \\ \midrule
            35                                   & \begin{tabular}[c]{@{}l@{}}SCHOLAR: (0.3361, 'thanhouser filmfare bollywood filmography directorial kumar telugu starred biopic hindi')\\ SCHOLAR+BAT: (0.5322, 'kumar bollywood directorial filmography telugu filmfare prasad malayalam bachchan hindi')\end{tabular}                              & 0.2888                                     \\ \midrule
            40                                   & \begin{tabular}[c]{@{}l@{}}SCHOLAR: (0.7394, 'batsman wicket bowled bowler bowling wisden cricketer selector inning crease')\\ SCHOLAR+BAT: (0.761, 'bowled wisden selector batsman bowler wicket cricketer crease spinner mcc')\end{tabular}                              & 0.3045                                     \\ \midrule
            41                                   & \begin{tabular}[c]{@{}l@{}}SCHOLAR: (0.4571, 'statute constitutionality plaintiff unconstitutional defendant judicial appellate amendment jurisdiction judiciary')\\ SCHOLAR+BAT: (0.4569, 'prosecutor prosecution investigator testified testimony conviction convicted verdict sentenced pleaded')\end{tabular}                              & 0.3137                                     \\ \midrule
            45                                   & \begin{tabular}[c]{@{}l@{}}SCHOLAR: (0.4178, 'edda mahabharata scripture purana goddess poem poetic shiva prose devotional')\\ SCHOLAR+BAT: (0.3379, 'northumbria inscription kingship deity shrine annals worshipped attested buddha vassal')\end{tabular}                              & 0.3658                                     \\ \midrule
            48                                   & \begin{tabular}[c]{@{}l@{}}SCHOLAR: (0.5286, 'cavalry grenadier flank bridgehead infantry bayonet brigade artillery regiment repulsed')\\ SCHOLAR+BAT: (0.3652, 'dso despatch raaf gallantry adjutant instructor aviator canberra airman citation')\end{tabular}                              & 0.544                                      \\ \bottomrule
        \end{tabular}}
    \caption{Fifteen aligned topic pairs from the Wiki dataset.}
    
    \label{tab:all_topic_pairs_wiki}
\end{table*}

\begin{table*}[]
    \centering
    \footnotesize
    \scalebox{0.77}{
        \begin{tabular}{@{}rlr@{}}
            \toprule
            \multicolumn{1}{c}{\textbf{Pair \#}} & \multicolumn{1}{c}{\textbf{\begin{tabular}[c]{@{}c@{}}SCHOLAR vs SCHOLAR+BAT\\ (NPMI, Top 10 Topic Words)\end{tabular}}} & \multicolumn{1}{c}{\textbf{JS Divergence}} \\ \midrule
            1                                    & \begin{tabular}[c]{@{}l@{}}SCHOLAR: (0.2333, 'scientist monster cgi alien creature scientists attack bullets aliens sci')\\ SCHOLAR+BAT: (0.2636, 'scientist alien creature monster aliens computer cgi space giant scientists')\end{tabular}                              & 0.0273                                     \\ \midrule
            4                                    & \begin{tabular}[c]{@{}l@{}}SCHOLAR: (0.165, 'vhs copy remember dvd ago tape saw video years loved')\\ SCHOLAR+BAT: (0.1844, 'vhs copy tape remember dvd bought ago saw video available')\end{tabular}                              & 0.0327                                     \\ \midrule
            8                                    & \begin{tabular}[c]{@{}l@{}}SCHOLAR: (0.1146, 'kids kid dad parents mom christmas decides dies santa guy')\\ SCHOLAR+BAT: (0.118, 'dad mom kids parents kid uncle decides christmas dies cat')\end{tabular}                              & 0.0379                                     \\ \midrule
            11                                   & \begin{tabular}[c]{@{}l@{}}SCHOLAR: (0.1968, 'adaptation version novel bbc versions jane kenneth handsome adaptations faithful')\\ SCHOLAR+BAT: (0.2181, 'adaptation novel book read books faithful bbc version versions novels')\end{tabular}                              & 0.0383                                     \\ \midrule
            15                                   & \begin{tabular}[c]{@{}l@{}}SCHOLAR: (0.2758, 'show episodes episode shows abc season aired sitcom television seasons')\\ SCHOLAR+BAT: (0.2678, 'seasons episodes show aired episode abc sitcom season television network')\end{tabular}                              & 0.0416                                     \\ \midrule
            17                                   & \begin{tabular}[c]{@{}l@{}}SCHOLAR: (0.0863, 'fails wooden lacks unconvincing shallow contrived wretched embarrassing thin embarrassment')\\ SCHOLAR+BAT: (0.1047, 'lacks pacing fails contrived flat irritating lacking chemistry unconvincing uninteresting')\end{tabular}                              & 0.0424                                     \\ \midrule
            22                                   & \begin{tabular}[c]{@{}l@{}}SCHOLAR: (0.174, 'documentary footage interviews music documentaries disc dvd musicians extras insight')\\ SCHOLAR+BAT: (0.1796, 'footage available documentary release dvd print interviews vhs subtitles audio')\end{tabular}                              & 0.0459                                     \\ \midrule
            27                                   & \begin{tabular}[c]{@{}l@{}}SCHOLAR: (0.1532, 'sheriff car town decides husband killer police investigate chase security')\\ SCHOLAR+BAT: (0.2504, 'murder murdered detective killer murderer police murders suspects secretary serial')\end{tabular}                              & 0.0531                                     \\ \midrule
            28                                   & \begin{tabular}[c]{@{}l@{}}SCHOLAR: (0.3054, 'christian religious god religion christ faith church jesus beliefs truth')\\ SCHOLAR+BAT: (0.1027, 'filmmaker intellectual filmmakers pretentious artistic subject content sake context claim')\end{tabular}                              & 0.0539                                     \\ \midrule
            31                                   & \begin{tabular}[c]{@{}l@{}}SCHOLAR: (0.1136, 'gags school rock band cartoons record boys principal radio metal')\\ SCHOLAR+BAT: (0.3144, 'songs musical singing sing dancing singer concert song numbers dance')\end{tabular}                              & 0.0556                                     \\ \midrule
            35                                   & \begin{tabular}[c]{@{}l@{}}SCHOLAR: (0.0813, 'development seemed boring predictable weak explanation slow potential interesting suspense')\\ SCHOLAR+BAT: (0.092, 'hour asleep minutes seemed sounded sat felt rented waste confusing')\end{tabular}                              & 0.0616                                     \\ \midrule
            40                                   & \begin{tabular}[c]{@{}l@{}}SCHOLAR: (0.1821, 'noir murder detective gritty crime cop thriller tough clint veteran')\\ SCHOLAR+BAT: (0.1027, 'cop dennis sheriff gangster boss agent villain hopper action chases')\end{tabular}                              & 0.0679                                     \\ \midrule
            41                                   & \begin{tabular}[c]{@{}l@{}}SCHOLAR: (0.095, 'porn cops girls random camera amateurish tedious amateur screaming chick')\\ SCHOLAR+BAT: (0.1548, 'kills killed killer screaming killing kill boyfriend woods walks dies')\end{tabular}                              & 0.0803                                     \\ \midrule
            45                                   & \begin{tabular}[c]{@{}l@{}}SCHOLAR: (0.1884, 'planet wars sci graphics space science game robot fiction weapons')\\ SCHOLAR+BAT: (0.0959, 'action development fighting sequences visuals realistic epic fight battles cool')\end{tabular}                              & 0.0925                                     \\ \midrule
            48                                   & \begin{tabular}[c]{@{}l@{}}SCHOLAR: (0.1732, 'book read books novel adaptation author reading disappointed adapted translation')\\ SCHOLAR+BAT: (0.1063, 'liked overall surprised disappointed enjoyed pleasantly pretty expectations seemed expecting')\end{tabular}                              & 0.1196                                     \\ \bottomrule
        \end{tabular}}
    \caption{Fifteen aligned topic pairs from the IMDB dataset.}
    
    \label{tab:all_topic_pairs_imdb}
\end{table*}

\end{document}